\documentclass[sigconf]{acmart}

\AtBeginDocument{%
  }

\setcopyright{acmlicensed}
\copyrightyear{2018}
\acmYear{2018}
\acmDOI{XXXXXXX.XXXXXXX}

\acmConference[Conference acronym 'XX]{Make sure to enter the correct
  conference title from your rights confirmation emai}{June 03--05,
  2018}{Woodstock, NY}
\acmISBN{978-1-4503-XXXX-X/18/06}

\usepackage{hyperref}
\usepackage{forest}

\usepackage{microtype}
\usepackage{subfigure}
\usepackage{graphicx}
\usepackage{color}
\usepackage{longtable}
\usepackage{pifont}
\usepackage{etoolbox}
\usepackage[font=small,labelfont=bf]{caption}

\usepackage{amsmath, amsfonts, amsthm, amssymb}
\usepackage[lined,boxed,commentsnumbered,ruled,vlined]{algorithm2e}
\usepackage{mathtools}
\usepackage{color}
\usepackage{adjustbox}
\usepackage{floatflt}
\usepackage{lipsum} 
\usepackage{xcolor}
\usepackage{tikz}
\usepackage{tabularray}

\usepackage{multirow}

\usepackage{threeparttable}
\usepackage{booktabs}    

\usepackage{makecell}        
\usepackage{caption}         

\newcommand{\fullcircle}{\tikz[baseline=-0.6ex]\node[draw, fill=black, circle, minimum size=1.6ex, inner sep=0pt] {};} 
\newcommand{\halfcircle}{%
\tikz[baseline=-0.6ex]{
    \node[draw, circle, minimum size=1.6ex, inner sep=0pt] {}; 
    \path[fill=black] (0,-0.8ex) arc[start angle=-90, end angle=90, radius=0.8ex] -- cycle; 
}} 
\newcommand{\emptycircle}{\tikz[baseline=-0.6ex]\node[draw, circle, minimum size=1.6ex, inner sep=0pt] {};} 

\definecolor{hidden-draw}{rgb}{0,0,0}
\definecolor{mycolor}{RGB}{248,233,204}

\newcommand{\datasetFont}{\text}
\newcommand{\ours}{\datasetFont{\texttt{Sim-to-Real}}\xspace}

\tikzstyle{my-box}=[
    rectangle,
    draw=hidden-draw,
    rounded corners,
    text opacity=1,
    minimum height=1.5em,
    minimum width=5em,
    inner sep=2pt,
    align=center,
    fill opacity=.5,
    line width=0.8pt,
]
\tikzset{
leaf/.style={
my-box,
minimum height=1.5em,
fill=mycolor, 
text=black,
align=left,
font=\footnotesize,
inner xsep=2pt,
inner ysep=4pt,
line width=0.8pt
}
}
\begin{document}

\title{A Survey of Sim-to-Real Methods in RL: Progress, Prospects and Challenges with Foundation Models}


\author{$^\dag$Longchao Da, $^\dag$Justin Turnau, $^\dag$Thirulogasankar Pranav Kutralingam, \\$^\ddag$Alvaro Velasquez, $^\dag$Paulo Shakarian, $^\dag$Hua Wei}
\affiliation{%
  \institution{$^\dag$Arizona State University, $^\ddag$DARPA}
  \country{}
}






\renewcommand{\shortauthors}{Da et al.}

\begin{abstract}

  Deep Reinforcement Learning (RL) has been explored and verified to be effective in solving decision-making tasks in various domains, such as robotics, transportation, recommender systems, etc. It learns from the interaction with environments and updates the policy using the collected experience. However, due to the limited real-world data and unbearable consequences of taking detrimental actions, the learning of RL policy is mainly restricted within the simulators. This practice guarantees safety in learning but introduces an inevitable sim-to-real gap in terms of deployment, thus causing degraded performance and risks in execution. There are attempts to solve the sim-to-real problems from different domains with various techniques, especially in this era of emerging techniques such as large foundation or language models that have cast light on the sim-to-real challenges and opportunities.
  This survey paper, to the best of our knowledge, is the first taxonomy that formally frames the sim-to-real techniques from key elements of the Markov Decision Process (State, Action, Transition, and Reward). Based on the framework, we cover comprehensive literature from the classic to the most advanced methods including the sim-to-real techniques empowered by foundation models, and we also discuss the specialties that are worth attention in different domains of sim-to-real problems. Then we summarize the formal evaluation process of sim-to-real performance with accessible code or benchmarks. The challenges and opportunities are also presented to encourage future exploration of this direction. We are actively maintaining a \href{https://github.com/LongchaoDa/AwesomeSim2Real.git}{\textcolor{red}{repository}} to include the most up-to-date sim-to-real research outcomes to help the researchers in their work~\footnote{\url{https://github.com/LongchaoDa/AwesomeSim2Real.git}}. 
\end{abstract}




\maketitle

\section{Introduction}

Reinforcement Learning (RL) algorithms are showing potential in multiple domains for their promising sequential decision-making abilities. In addition to gaming scenarios, the solutions are getting closer to real-world problems such as robotic control~\cite{kober2013reinforcement}, recommender systems~\cite{afsar2022reinforcement,chen2021survey}, healthcare~\cite{yu2021reinforcement,gottesman2019guidelines}, and transportation~\cite{haydari2020deep, wei2021recent}, among others.


Despite the frontier explorations of the RL-based methods, deploying the RL-learned policies in the real world is still challenging~\cite{chen2022towards,velasquez2023darpa}, especially in high-risk scenarios like autonomous driving~\cite{kiran2021deep} and disease diagnosis or chronic treatment~\cite{liu2017deep}. These real-world problems are struggling to benefit from the RL methods due to the gap between the simulator (used for policy learning) and reality (used for policy deployment), known as `\ours' gap.

\begin{table*}[htbp]
\setlength{\tabcolsep}{13pt}
\renewcommand{\arraystretch}{0.8}
\centering
\caption{Comparison with Existing Surveys on Sim-to-Real Survey Papers (Abbreviations: MDP = Markov Decision Process).}
\label{tab:compare-existing-works}
\resizebox{0.99\textwidth}{!}{%
\begin{tabular}{ccccccccc}
\toprule
& ~\cite{zhao2020sim} & ~\cite{pitkevich2024survey} & ~\cite{dimitropoulos2022brief} & ~\cite{zhu2021survey} & ~\cite{salvato2021crossing} & ~\cite{hu2023simulation} & ~\cite{miao2023parallel} & \textbf{Ours} 
                                                    \\ 
                                                    \midrule
\begin{tabular}[c]{@{}l@{}}\makecell{\vspace{-1mm} Methodological Taxonomy on MDP}\end{tabular}                         & \emptycircle & \emptycircle                 & \emptycircle                                  & \emptycircle                     & \emptycircle                               & \emptycircle                            & \emptycircle               & \fullcircle          \\ \midrule
\begin{tabular}[c]{@{}l@{}}\makecell{\vspace{-1mm}Application-Specific Analysis}\end{tabular}           & \halfcircle                & \fullcircle                 & \emptycircle                     & \halfcircle                    &  \halfcircle                     & \halfcircle                    & \halfcircle                    & \fullcircle             \\ \midrule
\begin{tabular}[c]{@{}l@{}}\makecell{\vspace{-1mm}Literature Review from RL Perspective}\end{tabular}               & \fullcircle            & \fullcircle            & \halfcircle                       & \fullcircle                    & \fullcircle                     & \halfcircle                     & \halfcircle                    & \fullcircle             \\ \midrule
\begin{tabular}[c]{@{}l@{}}\makecell{\vspace{-1mm}Methods Empowered by Foundation Models}\end{tabular} & \emptycircle                 & \emptycircle                    & \emptycircle                     & \emptycircle             & \emptycircle                   & \emptycircle                  & \emptycircle                         & \fullcircle             \\ \midrule
\begin{tabular}[c]{@{}l@{}}\makecell{\vspace{-1mm}Formal Sim-to-Real Evaluation Metrics}\end{tabular}                                                                          & \emptycircle                     & \fullcircle                        & \emptycircle                                 & \emptycircle                    & \emptycircle                  & \emptycircle                               & \halfcircle                         & \fullcircle             \\ \midrule
\begin{tabular}[c]{@{}l@{}}\makecell{\vspace{-1mm}Comprehensive Summary of Benchmarks}\end{tabular}                                     & \emptycircle                          & \fullcircle                             & \emptycircle                                 & \emptycircle                    & \halfcircle                              & \emptycircle                         & \halfcircle                  & \fullcircle            \\ \midrule
\begin{tabular}[c]{@{}l@{}}\makecell{\vspace{-1mm} Limitations Analysis}\end{tabular}                    & \fullcircle         & \halfcircle             & \emptycircle                & \halfcircle               & \emptycircle         &\emptycircle             & \emptycircle             & \fullcircle          \\ \midrule
\begin{tabular}[c]{@{}l@{}}\makecell{\vspace{-1mm}Future Research Direction}\end{tabular}         & \fullcircle        & \fullcircle     & \halfcircle            & \halfcircle         & \fullcircle             & \fullcircle            & \fullcircle                   & \fullcircle            \\ \bottomrule
\end{tabular}
}
\begin{threeparttable}
\begin{tablenotes}
\footnotesize
\item Wenshuai Zhao et al., 2020~\cite{zhao2020sim}; Andrei Pitkevich et al., 2024~\cite{pitkevich2024survey}; Konstantinos Dimitropoulos et al., 2022~\cite{dimitropoulos2022brief}; Wei Zhu, 2021~\cite{zhu2021survey}; Erica Salvato et al., 2021~\cite{salvato2021crossing}; \\ Xuemin Hu et al., 2023~\cite{hu2023simulation}; Qinghai Miao et al., 2024~\cite{miao2023parallel}.  \fullcircle: Discussed; \halfcircle: Partially discussed; \emptycircle: Not discussed.
\end{tablenotes}
\end{threeparttable}
\end{table*}

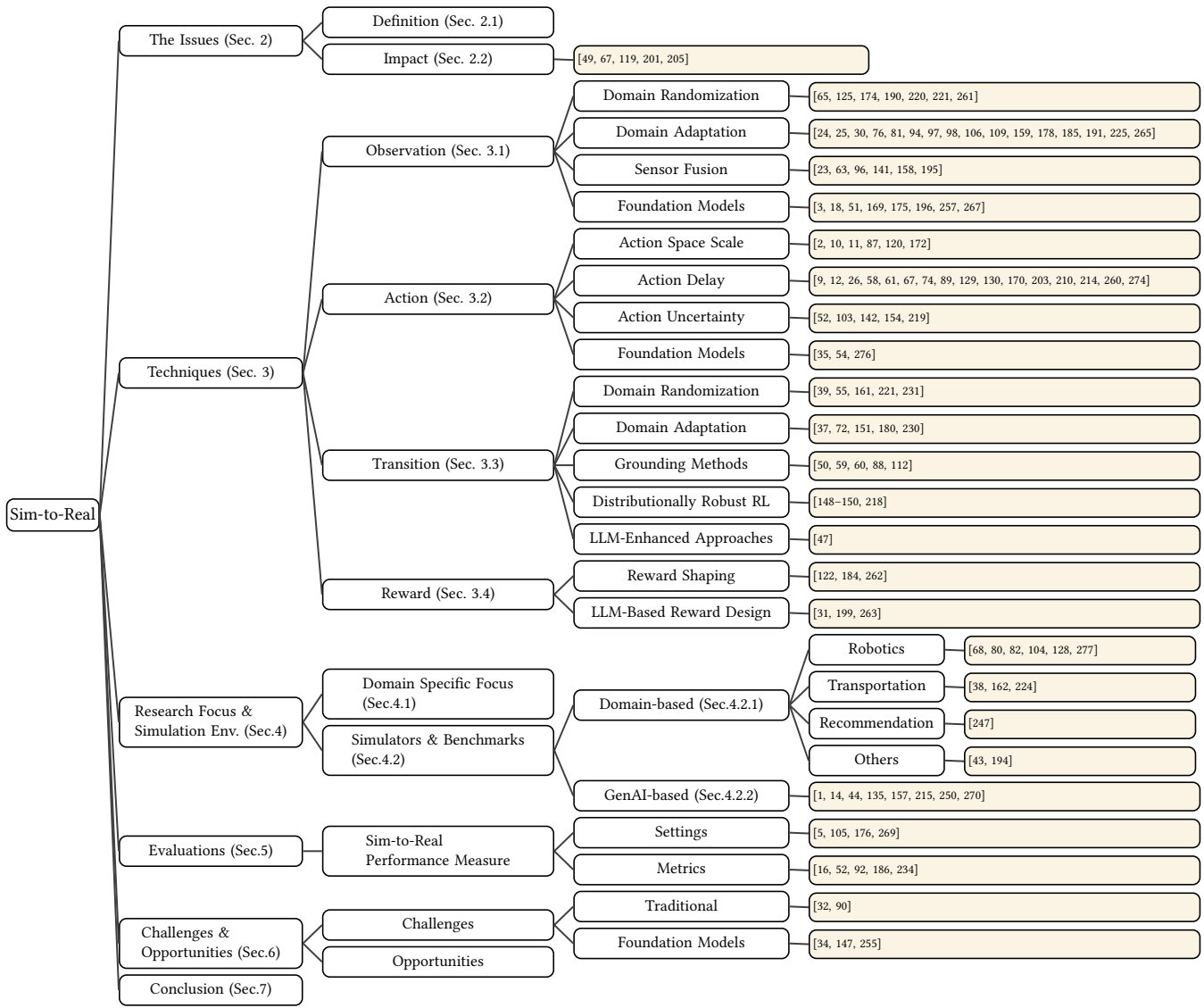
\begin{figure*}[t!]
    \centering
    \begin{adjustbox}{width=1\textwidth}
        \begin{forest}
            for tree={
                grow=east,
                reversed=true,
                anchor=base west,
                parent anchor=east,
                child anchor=west,
                base=center,
                font=\large,
                rectangle,
                draw=hidden-draw,
                rounded corners,
                align=left,
                text centered,
                minimum width=5em,
                edge+={darkgray, line width=1pt},
                s sep=3pt,
                inner xsep=2pt,
                inner ysep=3pt,
                line width=0.8pt,
                ver/.style={rotate=90, child anchor=north, parent anchor=south, anchor=center},
            },
            where level=1{text width=11em,font=\normalsize,}{},
            where level=2{text width=14em,font=\normalsize,}{},
            where level=3{text width=13em,font=\normalsize,}{},
            where level=4{text width=8em,font=\normalsize,}{},
            [
                Sim-to-Real, 
                [
                    {The Issues (Sec. \ref{sec:the_issue})},
                    [
                        Definition (Sec. \ref{sec:definition})
                    ],
                    [
                        Impact (Sec. \ref{sec:impact}),
                        [
                            \cite{da2023sim2real, dulac2019challenges, kormushev2013reinforcement, salvato2021crossing,schwall2020waymo}, leaf, text width=18em
                        ]
                    ]
                ],
                [
                    {Techniques (Sec. \ref{sec:tech})},
                    [
                        Observation (Sec. \ref{sec:obs}),
                        [
                            Domain Randomization,
                            [\cite{tobin2017domain, tiboni2023dropo, openai2019rubikscube, yuan2024maniwhere, mahesh2023bridging, helei2024challenging, pinto2017asymmetric}, leaf, text width=24em]
                        ],
                        [
                            Domain Adaptation,
                            [\cite{hu2022sim, carlson2019sensor, bousmalis2017unsupervised, ho2021retinagan, jing2023unsupervised, mahmood2018unsupervised, rao2020rl, park2021sim, jeong2020self, gu2020coupled, planamente2021da4event, bousmalis2018using, truong2021bi, zhang2019vr, gade2024domain, hoyer2023mic}, leaf, text width=24em]
                        ],
                        [
                            Sensor Fusion,
                            [\cite{mahajan2024quantifying, bohez2017sensor, reiher2020sim2real, hoglind2020lidar, liu2023world, ding2020longitudinal}, leaf, text width=24em]
                        ],
                        [
                            Foundation Models,
                            [\cite{achiam2023gpt, oquab2023dinov2, balazadeh2024synthetic, yu2024natural, zhao2024llm, nasution2024chatgpt, da2024segment, ren2024infiniteworld}, leaf, text width=24em]
                        ]
                    ],
                    [
                        Action (Sec. \ref{sec:act}),
                        [
                            Action Space Scale,
                            [\cite{anderson2021sim, abbas2024safety, alshiekh2018safe, odriozola2023fear, hamilton2023ablation, krantz2022sim}, leaf, text width=24em]
                        ],
                        [
                            Action Delay,
                            [\cite{dulac2019challenges, dezfouli2012habits, zhu2018hierarchical, sivakumar2019mvfst, harkavy2020utilizing, li2019reinforcement, li2020delay, al2020reinforcement, sun2023dynamic, firoiu2018human, antonova2017reinforcement, bouteiller2020reinforcement, yu2024dynamic, schuitema2010control, nath2021revisiting, derman2021acting}, leaf, text width=24em]
                        ],
                        [
                            Action Uncertainty,
                            [\cite{ilhan2021action, da2020uncertainty, lutjens2019safe, tessler2019action, liu2023efficient}, leaf, text width=24em]
                        ],
                        [
                            Foundation Models,
                            [
                                \cite{dalal2024local, zhu2020robosuite, chen2024rlingua}, leaf, text width=24em
                            ]
                        ]
                    ],
                    [
                        Transition (Sec. \ref{sec:tra}),
                        [
                            Domain Randomization,
                            [\cite{tobin2017domain, chen2021understanding, valassakis2020crossing, mehta2020active, dao2022sim}, leaf, text width=24em]
                        ],
                        [
                            Domain Adaptation,
                            [\cite{farahani2021brief, chen2020adversarial, tzeng2017adversarial, long2018conditional, pei2018multi}, leaf, text width=24em]
                        ],
                        [
                            Grounding Methods,
                            [\cite{hanna2017grounded, desai2020stochastic, karnan2020reinforced, desai2020imitation, da2023uncertainty}, leaf, text width=24em]
                        ],
                        [
                            Distributionally Robust RL,
                            [\cite{tang2024robust, liu2024upper, liu2025minimax, liu2024distributionally}, leaf, text width=24em]
                        ],
                        [
                            LLM-Enhanced Approaches,
                            [\cite{da2024prompt}, leaf, text width=24em]
                        ]
                    ],
                    [
                        Reward (Sec. \ref{sec:rew}),
                        [
                            Reward Shaping,
                            [\cite{place2023adaptive, zhang2024simple, kwon2023reward}, leaf, text width=24em]
                        ],
                        [
                            LLM-Based Reward Design,
                            [\cite{zhang2024accessing, chen2024evoprompting, ryu2024curricullm}, leaf, text width=24em]
                        ]
                    ]
                ],
                [
                    {Research Focus \& \\Simulation Env. (Sec.\ref{sec:domains})},
                    [ {Domain Specific Focus \\(Sec.\ref{sec:focus})},
                    ],
                    [ {Simulators \& Benchmarks \\ (Sec.\ref{sec:simulaor})},
                    [
                    {Domain-based (Sec.\ref{sec:domainsimu})},
                    [
                        Robotics,
                        [\cite{robustrl2024, gu2024humanoidgymreinforcementlearninghumanoid, li2024neuronsgym, james2019rlbenchrobotlearningbenchmark, ehsani2021manipulathor, zouitine2024rrls}, leaf, text width=14em]
                    ],
                    [
                        Transportation,
                        [\cite{mei2024libsignal, chen2024syntrac, tran2021tslib}, leaf, text width=14em]
                    ],
                    [
                        Recommendation,
                        [\cite{wu2021sim}, leaf, text width=14em]
                    ],
                    [
                        Others,
                        [\cite{crawley2001energyplus, Ray2019}, leaf, text width=14em]
                    ]
                    ]
                    [
                    {GenAI-based (Sec.\ref{sec:genaisimu})},
                    [\cite{cusumano2025robust, zhong2023guided, xu2023diffscene, li2023pac, ma2023learning, sun2020neupde, nvidia2023cosmos, Genesis}, leaf, text width=24em]
                    ]
                    ],
                ],
                [
                    {Evaluations (Sec.\ref{sec:eva})},
                    [
                        {Sim-to-Real \\ Performance Measure},
                        [ Settings,
                            [\cite{james2019sim, zhao2020sim, afzal2020study, osinski2020simulation}, leaf, text width=24em]
                        ],
                        [ Metrics,
                            [\cite{baek2024real, wagenmaker2024overcoming, polvara2020sim, da2020uncertainty, he2024bridging}, leaf, text width=24em]
                        ]
                    ]
                ],
                [
                    {Challenges \& \\ Opportunities (Sec.\ref{sec:challenge})},
                    [
                        Challenges,
                        [
                            Traditional,
                            [\cite{hawasly2022perspectives, chen2024trustworthy}, leaf, text width=24em]
                        ],
                        [
                            Foundation Models,
                            [\cite{yavas2023real, chen2024integrating, liu2023trustworthy}, leaf, text width=24em]
                        ]
                    ],
                    [
                        Opportunities
                    ]
                ],
                [
                    {Conclusion (Sec.\ref{sec:conclusion})}
                ]
            ]
        \end{forest}
    \end{adjustbox}
    \vspace{-4mm}
    \caption{Taxonomy of research on Sim-to-Real in RL that consists of the Issues, Techniques, Domain Discussion, and Evaluations.}
    \label{fig:tree}
\end{figure*}

The \ours gap is introduced in the policy training process and magnified in the deployment execution. Consequently, the well-trained RL policy suffers from severe real-world performance drop. In the worst case, there even exists a potential safety hazard given the unpredictable decisions under unseen scenarios. Some researchers attributed this to the transition gaps between the sim- and real- environments and proposed several aspects to tackle this gap from transition dynamics~\cite{valassakis2020crossing}, such as Domain Randomization and Domain Adaptation, etc. There was also literature discussing the gap introduced during the perception or execution period and proposed ways of Grounded Learning~\cite{sadeghi2018sim2real}. It is a pleasure to witness more and more attention cast on sim-to-rea RL, however, we find that different researchers are working on specific domains respectively~\cite{hofer2021sim2real, da2023sim2real, haiderbhai2024sim2real, wu2021sim}, and some great insights should be unified while specialties should be discussed in domain-specific aspects. Besides, with the explosive development of Large Foundation Models~\cite{liu2023pre, wei2022chain}, various effective approaches are proposed to integrate the foundation model's inference ability to downstream tasks~\cite{liu2022few}, and we observe its great potential to benefit sim-to-real transfer for RL methods. 

In this paper, we attempt to unify the commonly adopted, classic techniques from several sim-to-real domains, we provide a taxonomy that frames the majority of sim-to-real techniques based on the four elements of MDPs: Observation, Action, Transition, and Reward, and we also include the development of sim-to-real research from the classic to the most emerging techniques with foundation models. Then we discuss the challenges and solutions in domain-specific categories. After the introduction of the sim-to-real solutions in the training aspect, we categorize the evaluations into sim-to-real validation and Policy Evaluation, providing ways to effectively understand the policy performance.  

In summary, compared to existing survey papers as in Table.~\ref{tab:compare-existing-works}, our main contributions are \textbf{\ding{172} Taxonomy}: We propose a formal taxonomy for sim-to-real RL from issues, techniques, domains specialties and evaluations, and specifically, we categorize the technique solutions into the four foundation elements of MDPs; \textbf{\ding{173} Comprehensive Review}: We conduct comprehensive literature review that covers the most of related works, we reflect on the cause of sim-to-real gap and stem from classic, we introduce how large foundation models can benefit this research direction.  \textbf{\ding{174} Domain specific discussion}: Except for the techniques, we identify the unique challenges related to different real-world domains, and discuss potential prospects.

\section{The RL and Sim-to-Real Issue}\label{sec:the_issue}

In this section, we will start with a brief overview of the key concepts in Reinforcement Learning, including the MDP, and policy learning. And then, we will formally introduce the sim-to-real issue in RL. To avoid confusion, we include a notation explanation table to summarize key terms in Table~\ref{tab:notation}.

\begin{figure*}[t!]
    \centering
    \includegraphics[width=0.85\linewidth]{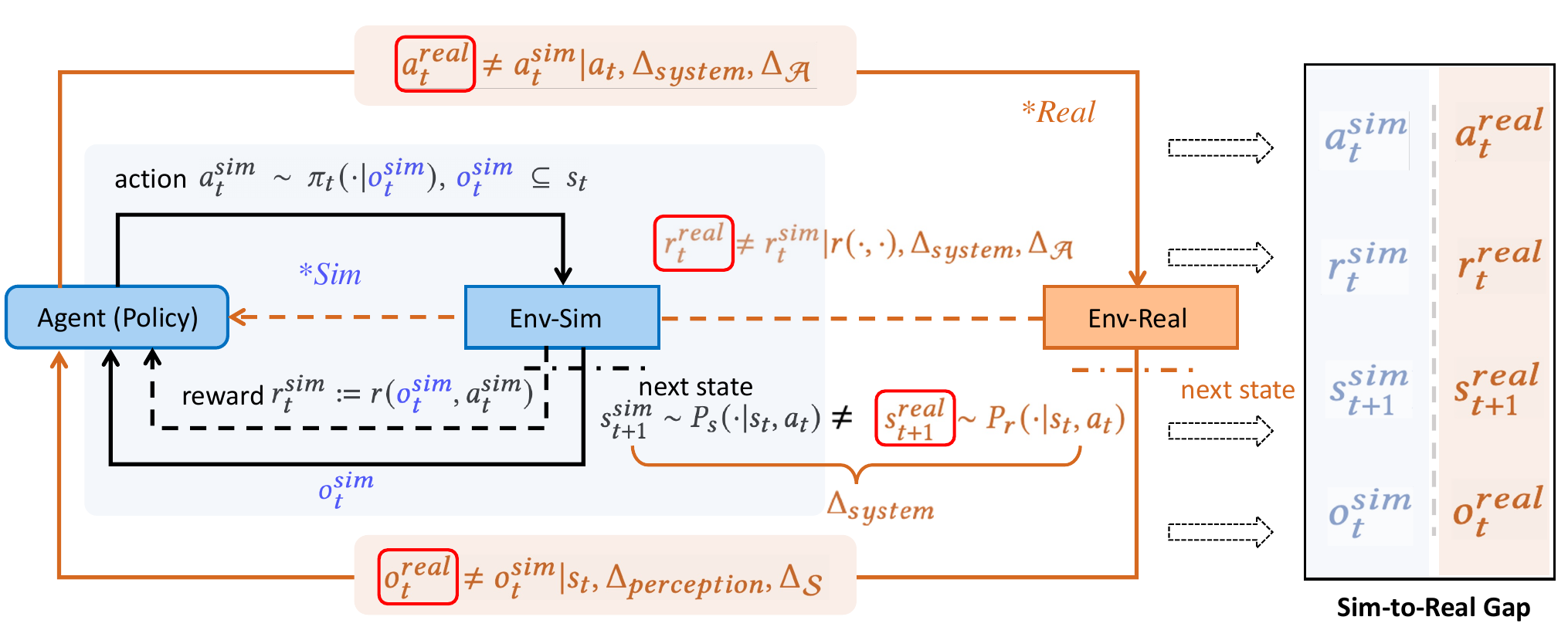}
    \caption{The overview of \ours issue causes. Four key sim-to-real (\textit{Sim2Real}) gaps in RL arise from discrepancies between the simulated environment (\textit{Env-Sim}) and the real-world environment (\textit{Env-Real}). The \textbf{Action Gap} ($a_t^{real} \neq a_t^{sim}$) originates from differences in system's mechanical state $\Delta_{system}$ or action space granularity $\Delta_{\mathcal{A}}$. The \textbf{Reward Gap} ($r_t^{real} \neq r_t^{sim}$) arises due to mismatches in the reward function between systems, and also the granularity of actions $\Delta_{\mathcal{A}}$. The \textbf{Next State Gap} ($s_{t+1}^{real} \neq s_{t+1}^{sim}$) reflects inaccuracies in the transition dynamics of the simulated environment $P_s(\cdot \mid s_t, a_t)$ compared to the real-world dynamics $P_r(\cdot \mid s_t, a_t)$. Lastly, the \textbf{Observation Gap} ($o_t^{real} \neq o_t^{sim}$) is from incomplete perception modules $\Delta_{perception}$ or the representations mismatch $\Delta_{\mathcal{S}}$. These collectively define the \ours challenge in RL.}
    \label{fig:main}
\end{figure*}

\subsection{Definition}\label{sec:definition}

\subsubsection{Reinforcement Learning (RL)}
Reinforcement Learning is a special machine-learning paradigm that empowers the learning of decision policy from the agent's interaction in an environment, the learning is directed by receiving feedback (reward) that comes along with the action. To maximize the accumulated reward, the policy is iteratively improved using various learning algorithms.

In general, the above RL learning procedure is often defined on a Markov Decision Process (MDP) $\mathcal{M}$ that satisfies formal mathematical modeling~\cite{feinberg2012handbook}, where  $\mathcal{M} = (\mathcal{S}, \mathcal{A}, \mathcal{T}, \mathcal{R}, \gamma)$: 

\begin{itemize}
    \item $\mathcal{S}$ is the state space that covers all possible situations that happen in the environment, either discrete or continuous.
    \item $\mathcal{A}$ is the action space that encompasses all possible actions an agent can take, either discrete or continuous.
    \item $\mathcal{T}$ is the transition function that defines the probability distribution of moving from state $s_t$ to state $s_{t+1}$ given that action $a_t$ is taken, i.e., $\mathcal{T} (s_{t+1}|s_t, a)$. The transition $\mathcal{T}$ is defined on $S \times A \times S \rightarrow \mathbb{R}$.
    \item $\mathcal{R}$: reward specifies the feedback an agent could obtain by taking action $a$ at state $s_t$ and moving to $s_{t+1}$. 
    \item $\gamma$ is the discount factor that determines the importance of future rewards.
    
\end{itemize}

The MDP starts with an initial state $\mu_0$ that is sampled from $\mathcal{S}$, an action $a_t$ will be output from current policy $\pi$ every time needs an action to further interact with the environment. The $\pi$ can be taken as a mapping from states to actions, $\pi : \mathcal{S} \rightarrow \mathcal{A}$, and the quality of $a$ is measured by value functions. To be concise, the RL's goal is to learn a policy that maximizes the accumulated expected return on the basis of rewards: 

\begin{equation}
    J(\pi) : = \mathbb{E}_{{(s, a)}\sim{\mu^{\pi}(s, a)}} [\sum_t{\gamma^t r_t(s_t, a_t, s_{t+1} \sim p(s_t, a_t))}]
\end{equation}
where $\mu^{\pi}(s, a)$ is the stationary state-action distribution under $\pi$, $a_t = \pi(s_t)$ and $p (s_t, a_t) = \mathcal{T} (s_{t+1}|s_t, a)$.

There are multiple branches of research on RL algorithms for faster and more stable learning processes, like value-based methods: Deep Q learning (DQN)~\cite{hester2018deep}, State-Action-Reward-State-Action (SARSA)~\cite{glascher2010states}, and policy gradient-based methods such as REINFORCE~\cite{sutton1999policy}, Proximal Policy Optimization (PPO)~\cite{schulman2017proximal}, and Deep Deterministic Policy Gradient (DDPG)~\cite{lillicrap2015continuous}, etc. This paper focuses on the sim-to-real problems and solutions, so it will not step into detail on the methods above. 

\subsubsection{Sim-to-Real of Reinforcement Learning}

In the context that the RL algorithm learns upon a MDP $\mathcal{M}$, the \ours issue can be described as the policy $\pi^i_s$ learned from $\mathcal{M}_s$ can not be well generalized to $\mathcal{M}_r$, where $\mathcal{M}_s$ represents the simulator environment $E_{sim}$ and $\mathcal{M}_r$ depicts the real world $E_{real}$. We formally define \ours gap as $G(\pi)$: 

\begin{equation}\label{eq:gap}
    G(\pi) : = \psi_s (\pi^i_s) - \psi_r (\pi^i_s) | \pi^i_s \sim \mathcal{M}_s, 
\end{equation}
where $\psi$ is any evaluation metric to quantify the performance of a policy, which should be calibrated and applied identically in the simulator $\psi_s$ and the real-world $\psi_r$ environment.

Such performance gap $G(\pi)$ is directly a result of policy interactions but is introduced in the policy learning process, and we can analyze the causes from the elements of   $\mathcal{M}_s$, which mainly encapsulates: $(\mathcal{S}_s, \mathcal{A}_s, \mathcal{T}_s, \mathcal{R}_s)$. Since the discount factor $\gamma$ is an ideal abstraction of future impact, which is inherently non-perfect, we will not discuss this factor in this survey paper.

Most of the decision-making is based on the accurate perception of the real world~\cite{ravlin1987effect}, if the observation exists mismatches between the environment for training (sim) and for applying (real), \ours gap arises. The observation gap arises for two reasons: \underline{1. The observed information's completeness - $\Delta_{perception}$}, most of the time the $o_t^{sim}$ is too perfect and ideal that $o_t^{sim} = s_t^{sim}$, while in $o_t^{real}$ exists missing information that $o_t^{real} \neq s_t^{real}$ -  partial observation problems (POMDP), so for rigorousness, we will use $o_t$ in this paper. \underline{2. The mismatch of feature representation - $\Delta_{\mathcal{S}}$}, caused by perception resolutions, sensor noises, etc., and then it leads to the difficulty of performing as expected in $E_{sim}$. 

Then, the action-taking also leads to \ours gaps in two ways: \underline{1. Action granularity - $\Delta_{\mathcal{A}}$}. The actions make real effects, such as grabbing and moving objects for robotics, but $a_t^{sim} \in \mathcal{A}_s$ are mostly oversimplified or discretized for simulator construction $\mathcal{M}_s$, and ideally executed in $E_{sim}$ by $a^{sim}_t \sim \pi_t (\cdot| \textcolor{black}{o}^{\textcolor{black}{sim}}_{\textcolor{black}{t}})$, $\textcolor{black}{o}^{\textcolor{black}{sim}}_{\textcolor{black}{t}} \subseteq s_t$. However, the control movement in real action space $a_t^{real} \in \mathcal{A}_r$ is essentially continuous and flexible, leading to meticulous control options in the real world.
\underline{2. System state gaps - $\Delta_{system}$}. During action execution, system latency is inevitable. Most simulators assume actions trigger instantly, but real-world mechanical components introduce delays, further aggravating the \ours gap.

Beside the observation and action making, the transition gap from environments is also a severe cause. We analyze this issue by discussing the intuitive `next-state divergence'. It explains a scenario that, given the same state $s_t$ and action $a_t$, for the next step state $s_{t+1}$ in given two environments $E_{sim}$ and $E_{real}$ are different, which is a result of the transition probability differences between $P_s$ and $P_r$ that: $P_s(s^{t+1}|s^t, a^t) \neq P_r(s^{t+1}|s^t, a^t)$, and the inherent cause is the system dynamics gap - $\Delta_{system}$ as shown in the Figure~\ref{fig:main}. The dynamic systems difference causes the challenge for policy learning and especially the deployment in $E_{real}$.  





The reward function $r^{sim}_t := r (\textcolor{black}{o}^{\textcolor{black}{sim}}_{\textcolor{black}{t}}, a^{sim}_t)$ is an another crucial aspect that an RL algorithms performs \ours gaps. There are two main reasons why an RL policy may perform unexpectedly due to the reward function: \underline{1. Reward function design in $\Delta_{system}$.} The reward function design is based on the understanding of behavior causes and expectations, but researchers tend to leverage accessible simulators to design rather than quantify the real world, then such system gaps would lead to in-comprehensive design of $\mathcal{R}_s$, such as uncovered real-world cases or unexpected actions, and further leads to performance impairment in $E_{real}$, even causes safety concerns. \underline{2. Cascade result of action delay or granularity difference $\Delta_{\mathcal{A}}$} also leads to undesired performance impairment since the reward is a direct consequence of action.

The above analysis conceptually covers the comprehensive causes of \ours problems in RL. In the following section, this paper will discuss the impact the \ours issue, and then focus on a categorized technical solution study based on the four aspects: Observation, Action, Transition and Reward. Since the Large Foundation Models are revolutionizing the major research areas, we also spend a section to introduce how each aspect has benefited or can potentially benefit from the the the foundation models\footnote{We take the LLM as a kind of Large Foundation Models.}.


\subsection{Impact}\label{sec:impact}
\textcolor{black}{
The \ours problem significantly impacts the usability of RL in the real world~\cite{dulac2021challenges}. This not only leads to monetary costs but also safety concerns, attracting significant attention across multiple research domains~\cite{dulac2019challenges, kormushev2013reinforcement}. In robotics, simulator-learned walking agents often fail in practical deployment~\cite{salvato2021crossing}, in autonomous-driving cars, fatal crashes happened in real-world executions~\cite{schwall2020waymo}, and in traffic signal control, the simulator-learned traffic light policy can hardly handle the realistic traffic dynamics, leading to unideal solutions~\cite{da2023sim2real}. These examples resulted in wasted training costs, time and limited real-world applicability. Thus, in this paper, we explore the causes of \ours deployment failure and categorize current solutions based on the components of the MDP, helping researchers easily identify issues and refer to relevant solutions.}

\section{Techniques}\label{sec:tech}
In this section, we formally introduce the solutions to \ours in main conceptions of MDP.

\subsection{Observation}\label{sec:obs}
\begin{figure*}[t!]
    \centering
    \includegraphics[width=0.9\linewidth]{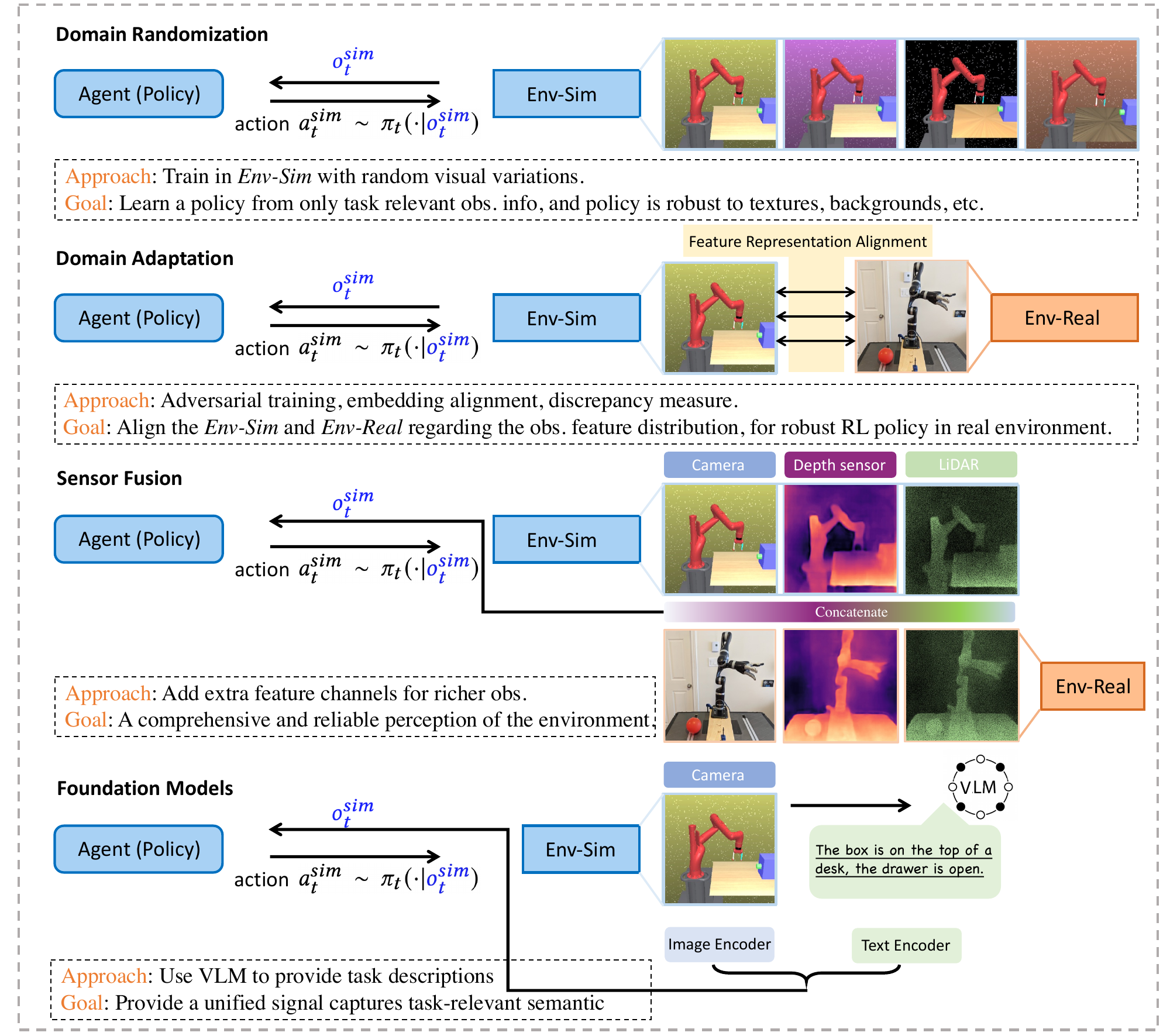}
    \caption{The four major types of the Sim-to-Real methods in Observation aspect using example from~\cite{mozifian2020intervention}. \textbf{Domain Randomization} enhances policy robustness by introducing a wide range of variations in simulated environments, enabling agents to generalize effectively to diverse real-world scenarios~\cite{tobin2017domain}. \textbf{Domain Adaptation} bridges the gap between simulated and real domains by aligning feature distributions, ensuring that policies trained in simulation perform consistently in real environments~\cite{tzeng2017adversarial}. \textbf{Sensor Fusion} integrates data from multiple sensors to provide comprehensive and reliable environmental perception~\cite{bohez2017sensor}, thereby compensating for the limitations of individual sensors, multiple observations provide a better grounding on the perception, thus mitigating the \ours issues. Lastly, \textbf{Foundation Models} increases the world depiction by leveraging the VLM to provide further task-level descriptions and encode such semantics info. into the agents' observations~\cite{yu2024natural}.}
    \label{fig:obs}
\end{figure*}

Bridging the sim-to-real (sim2real) gap in RL necessitates addressing discrepancies in observational data, particularly those arising from variations in sensor modalities such as cameras and tactile sensors. Various strategies have been developed to mitigate these differences as shown in Figre.~\ref{fig:obs}~\cite{mozifian2020intervention}:

\paragraph{\textbf{Domain Randomization.}} Observation-based domain randomization focuses on the state element 
\textit{S} of the Markov Decision Process (MDP) to mitigate the sim-to-real gap, distinguishing it from approaches like transitional domain randomization, which target transition dynamics as discussed in Section ~\ref{para:transitional_domain_randomization}. Observational domain randomization introduces variability into the visual parameters of the simulation—such as textures, lighting, and object positions—or into the sensors used for observations. By exposing models to a diverse range of simulated scenarios, this technique enhances robustness to the unpredictable nature of real-world environments ~\cite{tiboni2023dropo}. 

Common methods include randomizing textures, lighting conditions, object positions, object and background colors, as well as camera-related parameters such as position, depth, and field of view ~\cite{tobin2017domain, openai2019rubikscube, yuan2024maniwhere, mahesh2023bridging, helei2024challenging, pinto2017asymmetric}. In practice, these features are often used in tandem to create diverse training scenarios that encompass both subtle changes in appearance (e.g., lighting or colors) and more significant variations in spatial configuration or perspective (e.g., camera adjustments or object placements). For example, ~\cite{pinto2017asymmetric} demonstrates the effectiveness of domain randomization by training a robotic system to perform tasks such as pushing, moving, and picking up objects, even in the face of substantial changes in lighting, texture, and object position. By exposing the agent to a wide variety of simulated conditions, the system becomes more robust to real-world visual challenges, resulting in a more adaptable policy for use in deployment. A natural conclusion to draw would be to heavily randomize as many features as possible to produce as robust of an RL policy as possible, but this can destabilize RL policy training and lead to divergence ~\cite{yuan2024maniwhere}. Both ~\cite{yuan2024maniwhere} and ~\cite{openai2019rubikscube} address this by utilizing a curriculum-based domain randomization approach. ~\cite{openai2019rubikscube} introduces Automatic Domain Randomization (ADR), which both eliminates the need for manual fine-tuning of randomization ranges for domain randomization and uses a curriculum-based approach, gradually increasing environment difficulty as the policy improves. They utilized ADR to train both the policy and vision model, incorporating Gaussian noise with randomized parameters into observations during policy training, while randomizing visual features such as lighting conditions and camera perspectives to improve vision training. More recently, ~\cite{yuan2024maniwhere} introduced a generalizable framework for vision-based reinforcement learning which includes a curriculum-based approach to domain randomization for stabilizing RL training and improving sim-to-real transfer. Utilizing this curriculum-based approach, they achieved state-of-the-art performance for vision-based RL.

\paragraph{\textbf{Domain Adaptation.}} This category focuses on aligning the observation feature distributions between simulated and real data. The observation can be any sensible features that are helpful for decision-making, such as images~\cite{hu2022sim,you2019universal}, sensors~\cite{carlson2019sensor}, or LiDARs, etc. Techniques such as adversarial training~\cite{jing2023unsupervised, bousmalis2017unsupervised, mahmood2018unsupervised, rao2020rl, ho2021retinagan}, and embedding alignment~\cite{park2021sim} are employed to minimize the discrepancy between the two domains, enabling the model to perform consistently across both environments~\cite{hoyer2023mic}.

Going beyond the simple feature embeddings, \cite{jeong2020self} proposed a self-supervised framework to optimize latent state representation (e.g.,~\cite{rizzardo2023sim}) through sequence-based objectives, which demonstrates superior performance in visual robotic manipulation tasks. \cite{gu2020coupled} extended the concept of domain adaptation to real-synthetic depth data by coupling realistic degradation and enhancement techniques, it models realistic noise patterns in synthetic depth maps and enhances real-world depth data using a color-guided sub-network, achieving generalization to diverse real-world scenarios without requiring additional fine-tuning. And~\cite{planamente2021da4event} emphasizes the potential of domain adaptation across novel modalities by tackling the sim-to-real gap in event-based cameras.

However, those approaches require computationally expensive adaptation during training, the other branch of work considers the efficiency~\cite{bousmalis2018using}. In
~\cite{truong2021bi}, authors proposed Bi-directional Domain Adaptation to bridge the sim-vs-real gap in both directions: real2sim to bridge the visual domain gap, and sim2real to bridge the dynamics domain gap, which proves with dramatic speed-up compared to the traditional method. ~\cite{zhang2019vr} introduced a complementary real-to-sim adaptation framework, ``VR-Goggles'', that shifts the focus from adapting synthetic data to real domains to translating real-world image streams back into synthetic modalities during deployment, it minimizes computational overhead in the training phase while maintaining model performance across diverse real-world scenarios. Recently, ~\cite{gade2024domain} proposed an architecture that combines domain adaption and inherent inverse kinematics into one model, which helps reconstruct canonical simulation images from randomized inputs and improves robot grasping accuracy. A different line of work adapts ideas of metacognition from psychology~\cite{flavell1979metacognition} to AI systems~\cite{wei2024metacog}.  Here, a ``metacognitive model'' to identify or reason about failures in a base model.  Early work has focused on training an additional model to predict failures~\cite{daftry_introspective_2016,ramanagopal_failing_2018}, while more recent approach known as error detection rules allows for lightweight learned rules~\cite{kri24}.  These techniques have been applied to perception problems; integrating them in an RL framework is a promising future avenue to address the sim-to-real gap.

\paragraph{\textbf{Sensor Fusion.}} Combining data from multiple sensors can enhance the robustness of RL policies~\cite{mahajan2024quantifying}. For example, integrating visual data with depth sensors~\cite{bohez2017sensor}, combining LiDAR and camera inputs~\cite{reiher2020sim2real, hoglind2020lidar}, or merging auditory and inertial measurements allows the model to compensate for the limitations of individual sensors, leading to improved performance in complex tasks~\cite{liu2023world}. Specifically, ~\cite{ding2020longitudinal} proposed a multi-sensor fusion framework in four-wheel-independently-actuated electric vehicles, combining GPS and inertial measurements to address biases and noise in individual sensors, which highlights the critical role of sensor fusion in improving real-time estimations for sim-to-real applications, particularly in dynamic environments.

\paragraph{\textbf{Foundation Models.}} Recent advancements have explored the integration of large language models (LLMs)~\cite{achiam2023gpt} and multimodal foundation models~\cite{oquab2023dinov2} to understand physical world~\cite{balazadeh2024synthetic}, and further mitigate observation discrepancies in sim2real scenarios. These models, with their extensive pretrained knowledge and reasoning capabilities, can be leveraged to interpret and align observational data across domains. For example, natural language descriptions have been utilized to create a unifying signal that captures underlying task-relevant semantics~\cite{yu2024natural}, which are also known as semantic anchors, remain consistent in $E_{sim}$ and $E_{real}$, aiding in bridging the visual gap between simulation and reality~\cite{zhao2024llm}. Vision-Language Models (VLMs), by combining visual and textual data processing, can assist in generating descriptive annotations for sensory inputs, facilitating better understanding and alignment between simulated and real-world observations~\cite{nasution2024chatgpt, da2024segment}, and even facilitate simulation framework designs~\cite{ren2024infiniteworld}.

In summary, addressing observation discrepancies in sim2real transfer involves a combination of techniques aimed at enhancing model robustness and adaptability. The incorporation of LLMs presents a promising avenue for further reducing the observation gap, thereby improving the efficacy of RL policies in real-world applications.

\subsection{Action}\label{sec:act}

\begin{figure}[h!]
    \centering
    \includegraphics[width=0.9\linewidth]{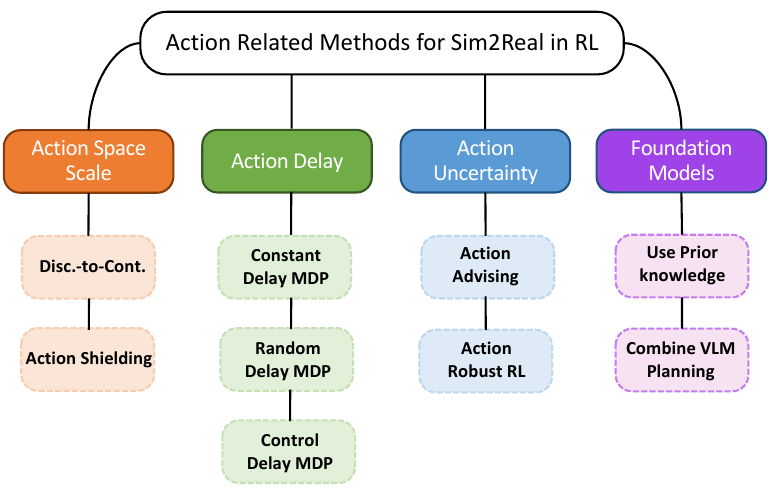}
    \caption{The taxonomy of action-related methods in sim2real RL.}
    \label{fig:actions}
\end{figure}

Action-taking is a key step to proceeding with any active control policy and results in the environment to make a difference. This section covers three main aspects of \texttt{action} that can mitigate the \ours problem, as shown in Figure~\ref{fig:actions}. The methods are categorized into \texttt{Action Space Scale}, \texttt{Action Delay}, and \texttt{Action Uncertainty}. 

\paragraph{\textbf{Action Space Scale}}
Actions have the most direct influence on the environment. However, due to the simulator's limitations, they are often discretized or simplified to reduce the design effort of fidelities. The most common scenario is the \underline{Discrete (sim) to Continuous} \underline{(real) gap}. To bridge such gaps between the high-level discrete action space learned by the agent and the robot’s real-world low-level continuous action space,~\cite{anderson2021sim,krantz2022sim} proposes a subgoal model to identify nearby waypoints in the simulator navigation graph during navigation tasks, which helps the policy under low fidelity to perform well in the real world. 
And \underline{Action Shielding}~\cite{abbas2024safety, alshiekh2018safe} focuses on ensuring that the actions selected by an agent in a simulated environment remain safe~\cite{odriozola2023fear}, feasible, and effective when transferred to the real world. Common methods often leverage a safety layer or filtering mechanism that evaluates an agent’s chosen actions against predefined constraints or real-world feasibility metrics~\cite{hamilton2023ablation}. For instance, during the transition from a discretized (sim) to continuous (real) action space, shielding can act as an intermediary, modifying or rejecting unsafe actions to prevent potential damage to the real-world system or environment. This ensures that high-level policies developed in the simulator can operate reliably and safely in dynamic, low-fidelity real-world scenarios.

\paragraph{\textbf{Action Delays}}
Another idealization of action-taking in the simulator is that the action often happens immediately. However, in the real world, it mostly comes with a delay~\cite{dulac2019challenges, dezfouli2012habits, zhu2018hierarchical}. Multiple domains tackle delayed action problems, such as network management~\cite{sivakumar2019mvfst, harkavy2020utilizing, li2019reinforcement, li2020delay}, which deals with the impracticality of real-time blocking or scheduling. In the energy domain,~\cite{al2020reinforcement, sun2023dynamic} manage energy without compromising the timely flow of data. Considering such delay variables in RL methods is an important step before real-world deployment.

In a \underline{Constant-delayed MDP} system,~\cite{firoiu2018human} proposes a predictive model inspired by how humans subconsciously anticipate the near future in physical environments to deal with delay-led consequences. In \underline{Random-Delay MDP} (RDMDP), early work~\cite{antonova2017reinforcement} empirically points out that randomized delays in the training process help to learn a more robust policy in the real world. Until~\cite{bouteiller2020reinforcement} formally defined the RDMDP and proposed a Delay-Correcting Actor-Critic (DCAC), which adopts action buffers and leverages delay measurements to correct for delays in the agent's actions. This approach generates actual on-policy sub-trajectories from off-policy samples, successfully improving the policies' performance under real-world action delay scenarios. In contrast,~\cite{yu2024dynamic} presents the Prediction model with Arbitrary Delay (PAD), a multi-step prediction model that mitigates cumulative error through a single prediction step rather than iterative updates, as in DCAC. PAD employs a gated unit that dynamically adjusts the feature extraction layers for different delays, enabling quick adaptation to random delay scenarios. Similarly,~\cite{schuitema2010control} defines the problem as a \underline{Control-delay MDP} and proposes two temporal difference-based methods, D-SARSA and D-Q, which compensate for action delays without state augmentation~\cite{nath2021revisiting} by updating Q-values based on effective delayed actions, improving performance under delayed conditions. The paper~\cite{derman2021acting} tackles execution delay in RL by using a Delayed-Q algorithm. Instead of relying on traditional state augmentation, which can exponentially increase complexity, this paper infers future states using a forward model based on the delayed action sequence. The algorithm then updates the Q-values with the inferred future state, allowing the agent to make more accurate decisions that compensate for action delays.

\paragraph{\textbf{Action Uncertainties}}
Action-taking inevitably involves uncertainty. Even a well-learned policy can encounter unseen scenarios, making real-world decision-making challenging. Incorporating uncertainty quantification brings great benefits for a simulator-trained policy to generalize to wider real-world scenarios. Here we cover two aspects of uncertainty-enhanced action taking: \texttt{Action Advising} and \texttt{Action Robust RL}.


\underline{Action Advising}~\cite{ilhan2021action}  is an RL technique where an agent receives guidance from a more experienced entity (e.g., a human or another agent) on which action to take in uncertain situations. Recently, the work~\cite{da2020uncertainty} proposes RCMP (Requesting Confidence-Moderated Policy advice), which uses epistemic uncertainty to guide action selection. RCMP estimates uncertainty by learning multiple value function estimates and computing their variance, providing a reliable measure of action confidence. This is especially useful for sim2real tasks where accurate decision-making under uncertainty is crucial. ~\cite{lutjens2019safe} introduces a Model Predictive Controller that prioritizes safer actions by evaluating each action's expected collision probability and uncertainty. Actions with lower uncertainty and lower collision probability are chosen, allowing the agent to cautiously avoid dynamic obstacles. This approach enables safer decision-making in safety-critical scenarios by avoiding high-risk actions when uncertainty is detected.

Another branch of research treats action-related uncertainty-aware Reinforcement Learning as \underline{Action Robust RL}. As a sub-domain of robust RL~\cite{wang2023model}, it is different from Action Advising, without relying on external advisors, it focuses on improving the robustness of an agent's actions in uncertain or adversarial environments with unexpected disruptions. In~\cite{tessler2019action}, the authors address action uncertainty by introducing two models: Probabilistic Action Robust MDP (PR-MDP) and Noisy Action Robust MDP (NR-MDP). 
These models help in selecting safer actions under uncertainty by considering adversarially affected outcomes, enabling the RL agent to maintain stable performance even under unexpected disturbances in real world. Following this work, the paper~\cite{liu2023efficient} introduces the ARRLC algorithm and handles action uncertainty by simulating the agent’s chosen action being replaced by an adversarial action with a probability $\rho$. ARRLC uses both optimistic and pessimistic estimates of the Q-function, allowing the agent to balance exploration and adversarial planning effectively.

\paragraph{\textbf{Foundation Models}}
Since foundation models are trained on massive corpus and show strong zero-shot capabilities, they are adopted to solve the generalizability challenges in unseen or rare scenarios' action-takings. Such as~\cite{dalal2024local} combines local policies with VLMs for motion planning, by training the simple local policies, these policies are serving as an action pool, e.g., \texttt{pick}, \texttt{open}, \texttt{close}, etc., and the foundation model will provide a planning strategy using these actions to finish the task. It shows superior performance on Robosuite benchmark~\cite{zhu2020robosuite}.  Similarly,~\cite{rajvanshi2024saynav} proposes SayNav, which grounds LLMs for dynamic planning to effectively navigate and finish tasks in new large-scale environments. Specifically, SayNav integrates an incremental scene graph generation, an LLM-based planner, and a low-level executor, and achieves state-of-the-art performance on the Multi-Object Navigation task.
\begin{figure}[h!]
    \centering
    \includegraphics[width=0.99\linewidth]{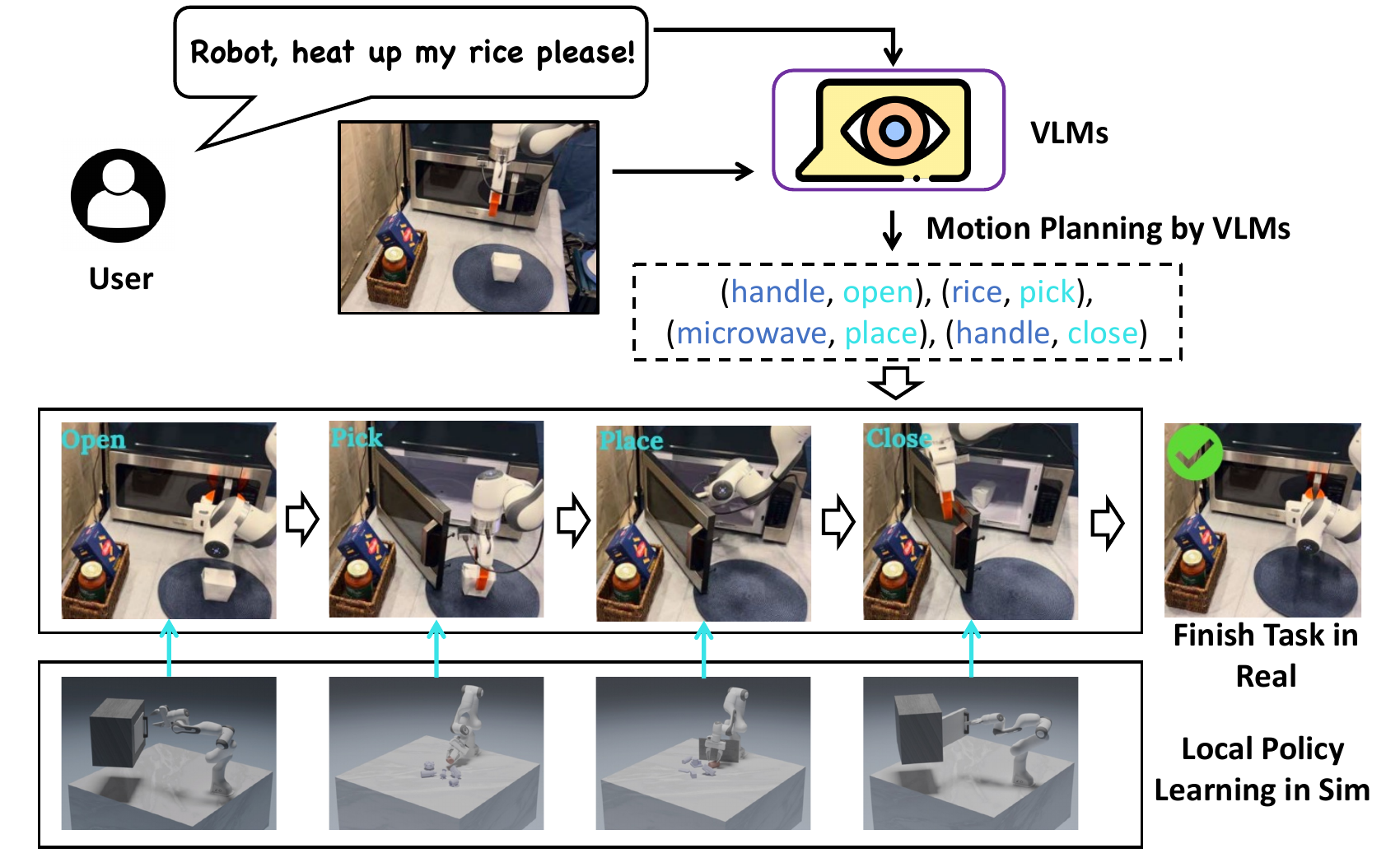}
    \caption{In \cite{dalal2024local}, a method is presented that uses the zero-shot capabilities of Vision Language Models (VLMs) to perform long-horizon manipulation tasks. Local policies are trained in the $E_{sim}$, while task execution occurs in the $E_{real}$, with the VLM coordinating the actions within the motion plans to achieve the task.}
    \label{fig:actionLLMs}
\end{figure}

The LLMs are also proved to be able to improve RL’s sample efficiency by leveraging their internal knowledge to generate preliminary rule-based controllers for robot tasks, which guide the exploration process and reduce the number of interaction samples required for effective learning~\cite{chen2024rlingua}.

\subsection{Transition}\label{sec:tra}

In \ours challenges, discrepancies in transition dynamics between simulated and real-world systems significantly impair policy deployment performance as showcased in exploration~\cite{da2023sim2real}, in this section, we will introduce five categories of methods that solve the \ours by bridging the transition dynamics gaps through policy learning.

\paragraph{\textbf{Domain Randomization}}\label{para:transitional_domain_randomization}
This method introduces variability within the simulator by randomizing physical parameters, enabling the simulated environment to encompass a wide range of potential real-world conditions~\cite{tobin2017domain, chen2021understanding,dao2022sim}. By exposing policies to diverse simulated scenarios, domain randomization enhances robustness, facilitating smoother transitions to real-world environments where conditions may not precisely match any single simulated setup. E.g., the work~\cite{valassakis2020crossing} randomized environmental factors such as friction, and motor torque, etc., ensuring that the trained policy is generalized effectively to real-world conditions without additional fine-tuning. And Mehta et al.~\cite{mehta2020active} introduced Active Domain Randomization (ADR), which enhances traditional domain randomization by concentrating training on the most challenging environment variations. ADR actively identifies and prioritizes configurations that lead to significant policy discrepancies, training agents on variations most likely to improve generalization to unseen real-world conditions.

\paragraph{\textbf{Domain Adaptation}}
This strategy focuses on aligning parameter distributions between simulated and real-world domains~\cite{farahani2021brief, chen2020adversarial}. This typically involves adversarial training techniques to minimize discrepancies between system features in both domains~\cite{tzeng2017adversarial, long2018conditional, pei2018multi}. By adapting the simulated domain to closely match the target domain's distribution, domain adaptation reduces mismatches in transition dynamics, enabling policies to generalize more effectively to real-world conditions while preserving specific traits essential for successful deployment.

\paragraph{\textbf{Grounding Methods}}
Grounding methods adjust simulator dynamics to align with real-world dynamics through grounded actions. Hanna and Stone~\cite{hanna2017grounded} proposed Grounded Action Transformation (GAT), which adjusts simulator dynamics to align with real-world dynamics through grounded actions. Building upon this, Desai et al.~\cite{desai2020stochastic} introduced Stochastic Grounded Action Transformation (SGAT), incorporating stochastic models into the grounding process using a probabilistic approach to model transition dynamics. Unlike GAT's deterministic setup, SGAT better approximates real-world stochastic behavior by learning a distribution over possible next states, enhancing robustness and policy transfer in variable environments. Karnan et al.~\cite{karnan2020reinforced} developed Reinforced Grounded Action Transformation (RGAT), which integrates RL directly into the grounding process. RGAT treats grounding as an RL problem, enabling end-to-end training of the action transformer as a single neural network, reducing error accumulation by learning a unified transformation function that optimally adjusts actions in simulation to match real-world dynamics. Further extending grounding methods, Desai et al.~\cite{desai2020imitation} introduced Generative Adversarial Reinforced Action Transformation (GARAT), framing action transformation as an Imitation from Observation (IfO) problem. GARAT employs a generative adversarial approach to minimize distribution mismatches between source (simulator) and target (real-world) dynamics. Through adversarial training, GARAT learns an action transformation policy that mimics the target environment by observing state transitions without explicit action labels, allowing for more accurate simulator adjustments. Additionally, Da et al.~\cite{da2023uncertainty} introduced uncertainty quantification to the GAT framework, enhancing decision-making reliability during \ours policy training.

\paragraph{\textbf{Distributionally Robust Learning}}
Recent works have tackled sim-to-real reinforcement learning by formulating it as a distributionally robust learning problem, the studies aim to design policies that can generalize to environments with unknown yet bounded transition shifts. In particular, ~\cite{liu2024distributionally} establishes provable efficiency guarantees for off-dynamics RL under linear function approximation, ensuring that policies trained in a source domain remain robust when deployed in a target domain with different dynamics. Building on this foundation,~\cite{liu2025minimax} introduces minimax optimal and computationally efficient algorithms for distributionally robust offline RL, further narrowing the gap between theoretical optimality and practical feasibility. Meanwhile,~\cite{tang2024robust} leverages linearly structured f-divergence regularization to enhance the robustness of offline RL methods by mitigating the impact of model uncertainty through carefully designed regularizers. Complementing these results,~\cite{liu2024upper} provides tight upper and lower bounds for distributionally robust off-dynamics RL, offering fundamental insights into the inherent performance limits under robustness constraints. Together, these contributions form a rigorous framework for distributionally robust learning, advancing both the theoretical understanding and practical applicability of sim-to-real RL in the face of significant distribution shifts.

\paragraph{\textbf{LLM-Enhanced Approaches}}
 Based on the grounding methods, a recent study~\cite{da2024prompt} tends to incorporate the LLM inference ability to the forward model's prediction of real-world dynamics, and then, based on better learned forward model, the predicted $\hat{s}_{t+1}$ is more reliable, so the inverse model can produce grounded actions in an effective way by taking the $(\hat{s}_{t+1}, s_{t})$. The empirical results show that such an attempt can efficiently improve the accuracy of real-world next-state prediction, and analysis shows a positive correlation between the accuracy of real-world dynamics depiction and the sim-to-real performance of the learned policy. This work offers a novel approach to incorporating LLMs' knowledge into \ours policy training.

In summary, addressing transition dynamics discrepancies in \ours involves a combination of traditional methods such as domain randomization, domain adaptation, and grounding methods, alongside emerging LLM-enhanced strategies. These approaches collectively enhance the robustness and adaptability of RL policies, facilitating more effective deployment in real-world applications.

\subsection{Reward}\label{sec:rew}

In reinforcement learning (RL), the design of reward functions is crucial for effective policy learning, especially when transferring from simulation to real-world environments (\ours). To address the challenges associated with reward functions in \ours scenarios, two primary categories of techniques have been explored: reward shaping and LLM-based reward design.

\paragraph{\textbf{Reward Shaping}}
Reward shaping techniques focus on modifying the reward or reward function to provide more informative feedback, thereby guiding the agent toward desired behaviors more efficiently. These methods are particularly beneficial in \ours contexts, where discrepancies between simulated and real environments can impede learning.

Potential-based reward shaping~\cite{badnava2023new} augments the original reward function with a potential function that reflects prior knowledge about the task. This ensures that the optimal policy remains unchanged while accelerating the learning process by providing intermediate rewards that guide the agent toward the goal~\cite{place2023adaptive, zhang2024simple}, automaton-guided reward shaping~\cite{velasquez2021dynamic, singireddy2023automaton} further refine this approach by leveraging structured representations - automata, to mitigate sparse reward challenges. It dynamically updates reward functions based on the utility of automaton transitions, improving both learning speed and robustness. Another approach involves assistant reward agents, which collaborate with the primary policy agent to generate supplementary reward signals based on future-oriented information. These auxiliary agents enhance sample efficiency and convergence stability by dynamically adapting the reward structure, facilitating effective exploration and exploitation during training~\cite{kwon2023reward}.

Another branch of work explores the possibility of augmenting the returns from a data-limited target environment, thus learning better-performed transferred policies: the work~\cite{guo2025off} introduces a reward augmentation technique based on trajectory distribution matching between the source (simulation) and target (real) environments, using imitation learning to transfer policies, while the work~\cite{wang2024return}  explores how decision transformers can be adapted for sim-to-real transfer via return augmentation, further improving return-conditioned supervised learning (RCSL) methods.

\paragraph{\textbf{LLM-Based Reward Design}}
The advent of large language models (LLMs) has opened new avenues for automating and refining reward function design in RL, particularly for complex tasks requiring nuanced reward structures. These techniques leverage LLMs' generative and reasoning capabilities to address reward design challenges in \ours scenarios.

Automated reward function generation uses LLMs to create reward function code from natural language task descriptions. Frameworks like CARD iteratively produce and refine reward functions without human intervention, aligning the generated rewards with task objectives through dynamic feedback mechanisms\textcolor{black}{~\cite{sun2024large}}. Evolutionary reward design combines LLMs with evolutionary algorithms to optimize reward functions. LLMs propose diverse candidate reward structures, which are then evaluated and improved through evolutionary search, resulting in more effective and tailored rewards\textcolor{black}{~\cite{narin2024evolutionary}}. Finally, text-to-reward frameworks such as Text2Reward~\cite{xie2023text2reward} automates the creation of dense reward functions from textual task specifications. By translating natural language descriptions into executable reward code, these systems reduce the need for domain-specific expertise while enabling rapid development of reward functions for various tasks~\cite{ryu2024curricullm}, and this reveals a potential angle in solving the \ours problems, especially in the zero-shot RL direction. Zero-shot or few-shot RL typically involves learning a representation that encapsulates task-relevant features. Based on this representation, policy generation occurs during inference time. At this critical stage, the design and refinement of reward functions play an important role. By dynamically adjusting rewards during the inference process, it is possible to effectively bridge the sim-to-real gap.

In summary, tackling the challenges of reward function design in \ours scenarios involves both traditional techniques like reward shaping and innovative approaches leveraging LLMs. These methodologies enhance the alignment between simulated training and real-world deployment, improving the robustness and effectiveness of RL policies.


\section{Research Focus and Simulation Environments}\label{sec:domains}
\ours transfer is a pervasive challenge across reinforcement learning (RL) applications, with each research domain adopting specialized simulators and benchmarks to address its unique real-world complexities. In this section, we categorize the literature into subdomains and provide an overview of the prevalent research resources—including simulation platforms and evaluation benchmarks, as well as the distinct research focuses in each domain.

\subsection{Domain-Specific Research Focus}\label{sec:focus}
\begin{table}[h!]
\centering
\caption{The Sim-to-Real research focus by different domains (an example on four categories)}
\label{tab:domain-focus}
\begin{tabular}{@{}c p{0.49\linewidth}@{}}
\toprule
\textbf{Domain Categorization} & \textbf{Research Focus} \\
\midrule
\multirow{3}{*}{\text{Robotics}} & \ding{172} Safety and risk mitigation \\
                                  & \ding{173} Multi-task capability \\
                                  & \ding{174} High accuracy \\[0.3ex]
\midrule
\multirow{3}{*}{\text{Transportation}} & \ding{172} Multi-agent coordination \\
                                          & \ding{173} Smooth transitions \\
                                          & \ding{174} Real-time decisions \\[0.3ex]
\midrule
\multirow{3}{*}{\text{Recommender}} & \ding{172} Large-scale online learning \\
                                      & \ding{173} Off-policy evaluation \\
                                      & \ding{174} Counterfactual sensitiveness \\[0.3ex]
\midrule
\multirow{3}{*}{\text{Others}} & \ding{172} Complex system dynamics \\
                                                 & \ding{173} Safety \& cost constraints \\
                                                 & \ding{174} Resource management \\[0.3ex]
\bottomrule
\end{tabular}
\end{table}

 For discussion, we choose the three most representative domains: Robotics, Transportation, Recommender Systems, and one general domain in sim-to-real research of reinforcement learning as in Table~\ref{tab:domain-focus}. In \textbf{Robotics}, the primary concerns are safety, multi-task capability, and high accuracy. Robotics applications often involve physical systems where any misstep can lead to safety hazards~\cite{abbas2024safety}, especially in sensitive contexts like medical procedures~\cite{haiderbhai2024sim2real}. Additionally, robotics demands versatility so that a single platform can handle varied tasks~\cite{fang2018multi}, and accuracy is critical when executing fine manipulation or precision-based tasks~\cite{akinola2020learning}. These factors drive researchers to develop solutions that ensure robust sensor integration and adaptive control strategies to bridge the gap between simulation and real-world performance.

In the \textbf{Transportation} domain, sim-to-real challenges are defined by the need for effective multi-agent coordination~\cite{chen2020toward}, smooth transitions in the traffic systems~\cite{balaji2010urban}, and real-time decision making~\cite{arel2010reinforcement}. Transportation systems, such as traffic signal control~\cite{li2015solving, wang2020large} or autonomous driving require policies that can handle rapidly changing environments and support interactions among numerous agents~\cite{yao2024comal}, and the training of decision policies requires traffic simulation to be well calibrated regarding the dynamic patterns (such as demand data)~\cite{zhou2014dtalite}. The emphasis on smooth transitions is particularly important as these systems often operate under mixed control scenarios where human expertise and automated processes must seamlessly cooperate. Moreover, the necessity for real-time responses further complicates the transfer from simulated environments, pushing researchers to focus on strategies that enhance responsiveness and stability in complex traffic networks.

In \textbf{Recommender Systems}, the sim-to-real gap emerges from the challenge of large-scale online learning to continuously update and refine recommendation policies~\cite{chen2021survey}, ensuring that they adapt to evolving user preferences. Robust evaluation methods such as off-policy evaluation~\cite{ma2020off, chen2022off, da2024probabilistic} and counterfactual analysis~\cite{li2022reinforcement} are also critical to quantifying and mitigating the discrepancies between simulated user models and actual user behavior. These efforts aim to develop RL-based recommendation strategies that maintain high performance and reliability when deployed in live environments.

In a broader context, the challenges extend to managing complex system dynamics~\cite{fang2023transferability}, ensuring safety and cost efficiency~\cite{li2022sim2real}, and handling resource management such as smart-grid energy control~\cite{tuli2022simtune, binyamin2022multi}. This category includes applications where the underlying processes are inherently uncertain and variable, making accurate simulation difficult. Research in this domain focuses on predictive control and robust optimization methods that capture the intricacies of real-world dynamics often simplified in simulations.

In summary, the \ours gap in reinforcement learning is deeply influenced by the unique demands of each application domain. These differences mean that RL methods must be specifically designed and tuned to address the nuances of each domain, which in turn shapes the strategies used to bridge the \ours transfer. Tailoring solutions in this way is important for achieving reliable and effective real-world performance.

\begin{table*}[t!]
\centering
\resizebox{\textwidth}{!}{ 
\begin{tblr}{
  column{3} = {c},
  column{7} = {c},
  column{8} = {c},
  cell{2}{1} = {r=18}{},
  cell{2}{2} = {r=12}{},
  cell{14}{2} = {r=7}{},
  cell{21}{1} = {r=17}{},
  cell{21}{2} = {r=14}{},
  cell{35}{2} = {r=3}{},
  cell{38}{1} = {r=7}{},
  cell{38}{2} = {r=4}{},
  cell{42}{2} = {r=3}{},
  cell{45}{1} = {r=6}{},
  cell{45}{2} = {r=4}{},
  cell{49}{2} = {r=2}{},
  cell{51}{1} = {r=2}{},
  vline{2} = {1}{}, 
  vline{2} = {2 - 50}{},
  hline{1-2, 14, 21, 35, 38, 42,45,49 ,51} = {-}{},
}
& \textbf{Category}  & \textbf{Name}   & \textbf{Year} & \textbf{Task}              & \textbf{Obs Type} & \textbf{\textcolor{black}{Action Type}} & \textbf{\textcolor{black}{License}} \\
\textbf{Robotics}        & Environment    & ORBIT~\cite{Mittal_2023}                & 2024          & Robot Learning             & Image \& Scalar             & Continuous     & BSD-3-Clause License  \\
                &                & CALVIN~\cite{mees2022calvinbenchmarklanguageconditionedpolicy} & 2022 & Robot Learning          & Image             & Continuous      & MIT License   \\
                &                & RoboSuite~\cite{zhu2020robosuite}       & 2020          & Robot Learning             & Scalar            & Continuous      & MIT License   \\
                &                & dm\_control~\cite{Tunyasuvunakool2020} & 2020          & Robot learning             & Image \& Scalar             & Continuous     & Apache-2.0 License   \\
                &                & SoftGym~\cite{lin2021softgymbenchmarkingdeepreinforcement} & 2020 & Robot Learning          & Image \& Scalar   & Discrete \& Continuous     & BSD-3-Clause License  \\
                &                & Assistive Gym~\cite{erickson2020assistive} & 2020        & Physics Simulation         & Image             & Continuous      & MIT License   \\
                &                & Meta-world ~\cite{yu2020meta} & 2019        & Meta-RL         & Image \& Scalar             & Continuous      & MIT License   \\
                &                & PyBullet~\cite{pybullet}                & 2017          & Physics Simulation         & Image             & Continuous      & zlib License   \\
                &                & RoboSumo~\cite{al2017continuous}        & 2017          & Multi-Agent RL             & Image             & Continuous      & N/A   \\
                &                & OpenAI Gym~\cite{openai_gym}            & 2016          & RL Algorithm Development   & Scalar            & Discrete \& Continuous     & MIT License   \\
                &                & MuJoCo~\cite{todorov2012mujoco}         & 2012          & Physics Simulation         & Scalar            & Continuous      & Apache-2.0 License   \\
                &                & Gazebo~\cite{koenig2004design}          & 2004          & Robotics Simulation        & Image             & Continuous      & Apache-2.0 License   \\
                  
                    & Sim-to-Real Benchmark & Robust Gymnasium~\cite{robustrl2024}         & 2024          & Robust Reinforcement Learning             & Image \& Scalar           & Discrete \& Continuous      & MIT License    \\
                    &                        & Humanoid-Gym~\cite{gu2024humanoidgymreinforcementlearninghumanoid} & 2024 & Humanoid RL             & Scalar             & Continuous      & N/A    \\
                     & & DISCOVERSE~\cite{discoverse2024} & 2024 & Robot Learning             & Image \& Scalar             & Continuous      & MIT License   \\
                    &                        & NeuronsGym~\cite{li2024neuronsgym}           & 2024          & Robot learning              & Image \& Scalar         & Continuous      & MIT License    \\
                    &                        & RRLS~\cite{zouitine2024rrls}                & 2023          & Multi-Agent RL             & Image             & Continuous      & MIT License    \\
                    &                        & ManipulaTHOR~\cite{ehsani2021manipulathor}   & 2021          & Robot learning              & Image \& Scalar             & Discrete      & MIT License    \\
                    &                        & RLBench~\cite{james2019rlbenchrobotlearningbenchmark} & 2019 & Robot learning              & Image             & Continuous      & RLBench-Licensed    \\

\textbf{Transportation}      & Environment & TorchDriveEnv~\cite{lavington2024torchdriveenvreinforcementlearningbenchmark} & 2024          & Autonomous Driving         & Image \& Scalar      & Continuous      & N/A   \\
                    &             & AutoVRL~\cite{sivashangaran2023autovrl}            & 2023          & Autonomous Navigation      & Image             & Continuous      & Apache-2.0 License    \\
                    &             & Waymax~\cite{gulino2023waymaxaccelerateddatadrivensimulator} & 2023          & Autonomous Driving         & Image \& Scalar      & Discrete \& Continuous      & Waymax License Agreement    \\
                    &             & MetaDrive~\cite{li2022metadrivecomposingdiversedriving}      & 2022          & Autonomous Driving         & Image              & Continuous      & Apache-2.0 License    \\
                   
                    &             & InterSim~\cite{sun2022intersiminteractivetrafficsimulation} & 2022          & Interactive Traffic Simulation & Image \& Scalar  & Continuous      & MIT License    \\
                    &             & TrafficSim~\cite{suo2021trafficsimlearningsimulaterealistic} & 2021          & Traffic Simulation         & Image \& Scalar      & Continuous      & N/A    \\
                    &             & SUMMIT~\cite{cai2020summitsimulatorurbandriving}           & 2020          & Autonomous Driving         & Image \& Scalar      & Discrete \& Continuous      & MIT License   \\
                    &             & SMARTS~\cite{zhou2020smartsscalablemultiagentreinforcement} & 2020          & Autonomous Driving         & Image \& Scalar      & Discrete \& Continuous      & MIT License    \\
                    &             & deepdrive-zero~\cite{craig_quiter_2020_3871907}            & 2020          & Autonomous Driving         & Image              & Continuous      & MIT License    \\
                    &             & CityFlow~\cite{Zhang_2019}                  & 2019          & Traffic Signal Control     & Image \& Scalar      & Discrete      & Apache-2.0 License    \\
                    &             & highway-env~\cite{Leurent_An_Environment_for_2018}         & 2018          & Autonomous Driving         & Image \& Scalar      & Discrete \& Continuous      & MIT License   \\
                    &             & SUMO~\cite{lopez2018microscopic}        & 2018          & Traffic Simulation         & Scalar            & Discrete      & EPL-2.0 License    \\
                    &             & CARLA~\cite{dosovitskiy2017carla}                   & 2017          & Autonomous Driving         & Image              & Continuous      & MIT License    \\
                     &             & Duckie-MAAD~\cite{candela2022transferringmultiagentreinforcementlearning} & 2017 & Multi-Agent RL, Autonomous Driving & Image \& Scalar & Discrete      & N/A   \\

                    & Sim-to-Real Benchmark& SynTraC~\cite{chen2024syntrac}          & 2024          & Traffic Signal Control      & Image             & Discrete      & N/A    \\
                    & & LibSignal~\cite{mei2024libsignal}          & 2023          & Traffic Signal Control      & Scalar             & Discrete      & N/A   \\
                    & & TSLib~\cite{tran2021tslib}          & 2021          & Traffic Signal Control      & Image \& Scalar             & Discrete      & 0BSD License    \\
\textbf{Recommender Systems} & Environment& RecSim~\cite{ie2019recsim}           & 2019          & User Behavior Simulation   & Scalar            & Discrete      & Apache-2.0 license    \\
                        & & SlateQ~\cite{ie2019reinforcementlearningslatebasedrecommender}          & 2019          & RL Environment for Recommendation     & Scalar             & Discrete      & N/A    \\
                        & & RecoGym~\cite{recogym}          & 2018          & RL Environment for Recommendation      & Scalar             & Discrete      & Apache-2.0 license    \\
                    & & Virtual-Taobao~\cite{shi2018virtualtaobaovirtualizingrealworldonline}          & 2018          & RL Environment for Retail Simulation      & Scalar             & Discrete      & N/A    \\
                    
                   & Sim-to-Real Benchmark & KuaiSim~\cite{zhao2023kuai}          & 2023          & User Behavior Simulation      & Scalar             & Discrete      & Apache-2.0 license   \\
                    &  & S2R-Rec~\cite{wu2021sim}          & 2021          & Interactive Recommendation & Scalar   
                   & Discrete     & N/A    \\
                        & & RL4RS~\cite{wang2023rl4rsrealworlddatasetreinforcement}          & 2021          & RL Environment for Recommendation      & Scalar             & Discrete \& Continuous      & CC-BY-SA-4.0 license    \\
\textbf{Others}             & Environment & OpenAI Gym Retro~\cite{openai_gym_retro} & 2018          & Video Game Simulation      & Image             & Discrete     & MIT License    \\
                 
                 & & AI2-THOR\cite{kolve2022ai2thorinteractive3denvironment}          & 2017          & Navigation, Interaction    & Image             & Discrete \& Continuous     & Apache-2.0 license   \\
                 & & DeepMind Lab~\cite{beattie2016deepmindlab}          & 2016         & Interactions     & Image        & Discrete     & GNU General Public License    \\
                 & & Arcade Learning Environment (ALE)~\cite{Bellemare_2013}          & 2013          & Video Game Simulation     & Image             & Discrete      & GPL-2.0 license   \\
                & Sim-to-Real Benchmark & Safety Gym~\cite{Ray2019}          & 2019          & Safe RL     & Image \& Scalar  & Continuous      & MIT License    \\        
                 & & EnergyPlus~\cite{crawley2001energyplus}       & 2001          & Building Energy Simulation & Scalar            & Discrete \& Continuous      & BSD 3-Clause License    \\
                
\end{tblr}
}
\caption{Summary of Simulators and Benchmarks Across Various Fields. In each field, we categorize into plain \textbf{Environments} and \textbf{Sim-to-Real Benchmark} which is specially designed to support Sim-to-Real tasks.}
\label{tab:simulators_benchmarks}
\end{table*}

\subsection{Simulators and Sim-to-Real Benchmarks 
}\label{sec:simulaor}
In this section, we introduce the recent simulation platforms and sim-to-real benchmarks to bring convenience to researchers for their study. According to the discussion in Sec.~\ref{sec:focus}, we first introduce the `Simulators' and \ours `benchmarks' from a domain-specific level, which is summarized in Table.~\ref{tab:simulators_benchmarks}. Second, we introduce how generative AI methods enhance sim-to-real through simulations such as logic-integration, physics-augmentation, and world models, etc. 

\subsubsection{\textbf{Domain Specific Simulator and Benchmark}}\label{sec:domainsimu}

\paragraph{\textbf{Robotics}}
There are various simulators developed to model robotic systems with high fidelity, there are also researchers who tend to leverage LLMs with multi-modal and reasoning
capabilities for complex and realistic simulation task creation~\cite{hua2024gensim2, katara2023gen2sim}, thus enabling the policy training in simulator scenarios closer to what may be encountered in the real world.

Gazebo~\cite{koenig2004design} is an open-source 3D robotics simulator that integrates with the Robot Operating System (ROS) to provide realistic rendering of environments and physics for testing robot models and algorithms. MuJoCo~\cite{todorov2012mujoco} (Multi-Joint dynamics with Contact) is a physics engine designed for research in robotics and biomechanics, offering fast and accurate simulation of complex dynamical systems. PyBullet~\cite{pybullet} serves as an easy-to-use Python module for physics simulation, supporting both robotics and RL research with real-time collision detection and multi-body dynamics. OpenAI Gym~\cite{openai_gym_retro} is a widely-used toolkit for developing and comparing RL algorithms, featuring a variety of environments, including robotic simulations, to standardize the evaluation process.

To evaluate the transfer of learning from simulation to the real world in robotics, specific benchmarks are often utilized. RoboSuite~\cite{zhu2020robosuite} is a simulation framework designed for robot learning, providing a collection of benchmark tasks and environments to test the performance of RL algorithms in manipulation and control tasks. OpenAI’s RoboSumo offers a virtual environment where humanoid robots compete in sumo wrestling, serving as a valuable platform for multi-agent RL research and sim-to-real transfer studies. The Meta-World~\cite{yu2020meta} focuses on the evaluation of meta-RL algorithms on task distribution levels to enable the goal of generalization to new behaviors.

\paragraph{\textbf{Transportation}}
Simulators are employed to model complex traffic systems and train RL agents for tasks such as traffic signal control, autonomous driving, and fleet management. SUMO~\cite{lopez2018microscopic} (Simulation of Urban MObility) is an open-source, portable traffic simulation package capable of handling large road networks. CARLA~\cite{dosovitskiy2017carla} (Car Learning to Act) is another open-source simulator developed for autonomous driving research, featuring realistic urban environments and support for sensor suites, enabling the development and validation of autonomous driving systems. AutoVRL~\cite{bantel2024high} is a high-fidelity autonomous ground vehicle simulator built on the Bullet physics engine, specifically designed for sim-to-real deep RL in autonomous navigation tasks. Benchmarks in transportation often involve standardized scenarios within these simulators to evaluate the performance of RL algorithms in managing traffic flow, reducing congestion, and enhancing safety.

\paragraph{\textbf{Recommender Systems}}
Simulators are used to model user interactions and preferences, allowing RL agents to learn optimal recommendation strategies in controlled environments. RecSim~\cite{ie2019recsim} is a configurable platform designed to simulate user behavior in recommender systems, enabling the study of RL algorithms. S2R-Rec~\cite{wu2021sim} addresses the challenges of transferring policies from simulated environments to real-world settings in interactive recommendation systems, employing off-dynamics RL to bridge the gap between simulation and reality. Benchmarks in this area typically assess RL strategies based on metrics such as user engagement, satisfaction, and retention, using both simulated and real-world datasets.

\paragraph{\textbf{Other Domains and Alternatives}}
Simulators also extend to domains such as finance, healthcare, and energy management, where specialized simulators and benchmarks are tailored to address unique challenges. OpenAI Gym Retro is a platform designed for RL research on video games, supporting the study of generalization and transfer learning across various game environments. EnergyPlus is a simulation program for modeling building energy consumption, enabling RL research focused on optimizing energy management systems.  A different way to view simulators is to have a framework that allows for the creation of arbitrary scenarios.  Along these lines, \cite{mukherji2024semantic} leverages temporal logic programming~\cite{dekhtyar1999temporal,shakarian2011temporal} to create a ``semantic proxy'' of a real or simulated environment.  Here, a temporal logic program is used to replace the simulator, which has the advantage of not requiring the Markov assumption.  The authors show that an RL agent trained in the semantic proxy performs comparably in the simulation.  This may have implicaitons for sim-to-real translation going forward.

In summary, the selection of appropriate simulators and benchmarks is important for sim-to-real research in reinforcement learning. By categorizing these tools according to application domains, researchers can focus on the most relevant resources, ensuring a structured and systematic approach to developing and deploying RL methodologies in diverse real-world scenarios.

\subsubsection{\textbf{GenAI-based Simulations}}\label{sec:genaisimu}

There is a trend in simulation research for sim-to-real that involves Generative AI (GenAI) methods. These approaches adopt large-scale generative models (or foundation models) in simulation contexts, aiming to produce synthetic environments, physics dynamics, or training data that are both highly realistic and diverse. 

\paragraph{\textbf{Brute-Force Scaling}}
Large-scale simulation frameworks backed by massive computing and data can expose RL agents to billions of samples, thus empowering RL policies with unexpected generalizability. For instance, in work~\cite{cusumano2025robust}, researchers scaled simulation-based training to 1.6 billion km of self-play generated driving data, observing “emergent” realistic and robust behaviors during testing. These results suggest that sheer scale can produce highly capable policies.

\paragraph{\textbf{Logic-Integrated Simulation}}
Beyond straightforward `scaling law'-based generalizability, there is growing interest in incorporating formal logic or constraints directly into simulation. For example, frameworks have used Signal Temporal Logic (STL) to specify high-level requirements like traffic rules~\cite{zhong2023guided} or safety constraints~\cite{xu2023diffscene} and then automatically generate traffic scenarios that satisfy these conditions. Such logic-augmented simulations enable RL practitioners to produce training data tailored to specific operational or ethical constraints (e.g., minimal collisions, compliance with road rules), and then reduce the sim-to-real gap in critical domains like autonomous driving.

\paragraph{\textbf{Differentiable \& Physics-Augmented Simulations}}
Recent work on differentiable simulators integrates physics equations directly into neural networks, enabling gradients to flow through environment dynamics. This allows for more precise fine-tuning of simulation parameters against real-world observations.
PAC-NeRF~\cite{li2023pac} and other neural PDEs (partial differential equation) solvers~\cite{sun2020neupde, ma2023learning} can learn material or fluid properties from motion data. They can then simulate complex deformable or fluid interactions with minimal human-engineered approximations, providing a rich training ground for RL agents.

\paragraph{\textbf{Generative World Models and Physics Engines}} Large-scale generative models can also function as comprehensive world models or physics engines. For instance, NVIDIA COSMOS~\cite{nvidia2023cosmos} provides a unified “world foundation model” capable of coordinating multiple simulation elements—such as scene composition, lighting, and object configurations—at scale, greatly enriching the variety of training scenarios. Similarly, Genesis~\cite{Genesis} positions itself as a “Generative and Universal Physics Engine,” employing data-driven methods to capture a broad range of physical behaviors. These approaches allow for highly diverse and physically grounded simulations that move beyond traditional, hand-engineered routines and thus provide much richer training interactions for policy learning.

\section{Evaluation of Sim-to-Real}\label{sec:eva}

Evaluating sim-to-real (\ours) transfer in reinforcement learning (RL) is very important for understanding how well policies trained in simulation perform in real-world environments, it is of practical use when actually considering the deployment of RL policies in the real world. This evaluation typically involves three primary settings: sim-to-real, sim-to-scale-down-real, and sim-to-sim, each with distinct considerations and benefits, and detailed evaluation design is closely related to the domain tasks, the metrics vary based on the domain considerations.

\begin{figure}[h!]
    \centering
    \includegraphics[width=0.99\linewidth]{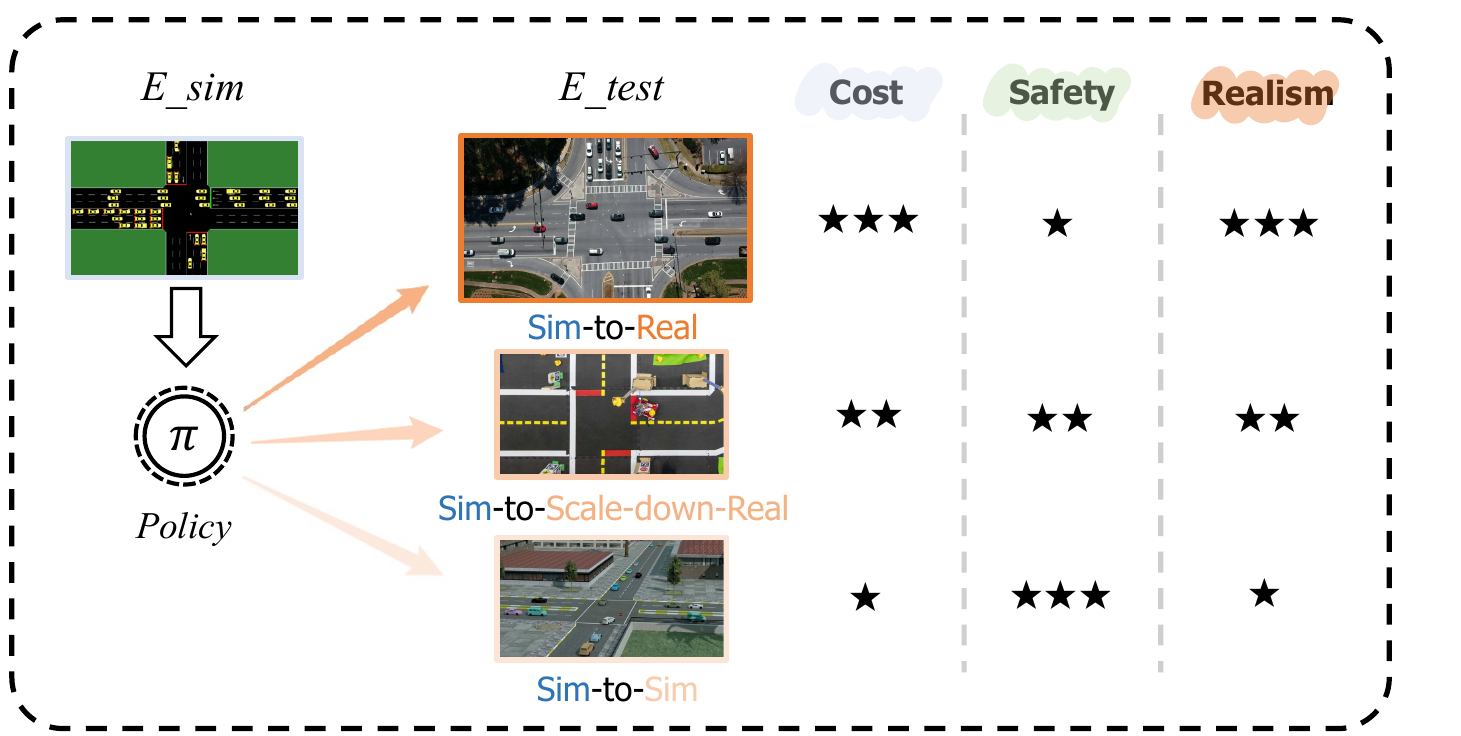}
    \caption{Comparison of three types of evaluation settings: Sim-to-Real, Sim-to-Scale-down-Real, and Sim-to-Sim, based on cost, safety, and realism. Sim-to-Real offers the highest realism but at a higher cost and lower safety. Sim-to-Scale-down-Real~\cite{paull2017duckietown} balances safety and cost by using controlled environments while maintaining moderate realism. Sim-to-Sim (from low-fidelity to high-fidelity) is the most cost-effective and safe option but might sacrifice realism due to the simulated nature of the environment.}
    \label{fig:evaluations}
\end{figure}

\subsection{Settings}
In this section, we will introduce three common ways of conducting the evaluation for \ours methods, and provide a comparison in three dimensions covering cost, safety, and realism as in Figure.~\ref{fig:evaluations}.

\textbf{Sim-to-Real Setting:} Sim-to-real evaluations involve deploying policies trained in simulation directly onto real-world physical systems. This setting is essential for domains requiring real-world interaction, such as robotics and autonomous vehicles. It can receive real-time actual feedback from the real environment but is not an ideal strategy for most of the verification experiments due to the unexpected behaviors of learned policies (especially for neural networks).

\textbf{Sim-to-Scale-down-Real Setting:} Due to the safety, cost, and fault-tolerance concerns, most of the evaluation configs do not happen in the real world directly, instead, specialized, scale-down testbeds are designed to facilitate these evaluations, ensuring safety and reliability. For example, in robotics, physical test environments equipped with motion capture systems and safety measures allow for controlled testing of robotic policies~\cite{afzal2020study}. Similarly, autonomous vehicle testing may utilize closed tracks that replicate real-world driving conditions to evaluate the simulated training~\cite{osinski2020simulation}.

\textbf{Sim-to-Sim Setting:} Due to the high costs and practical challenges associated with real-world testing, sim-to-sim evaluations are commonly employed as a preliminary step. In this setting, policies trained in one simulated environment are tested in a different, often more realistic or varied, simulation. This approach allows researchers to assess the robustness and generalization capabilities of RL agents under diverse conditions without the expenses and risks of real-world deployment. For instance, varying the physics parameters, sensor noise, or environmental dynamics between training and testing simulations can provide insights into how well a policy might transfer to reality~\cite{james2019sim}.

Selecting appropriate evaluation settings and metrics is vital for accurately assessing sim-to-real transfer in reinforcement learning. Sim-to-sim evaluations, where the agent is evaluated across different simulated environments, offer a cost-effective means to gauge potential real-world performance. Sim-to-real evaluations, on the other hand, directly assess the agent's performance in practical scenarios, providing definitive insights into its applicability. Combining these evaluation settings with robust metrics enables a comprehensive assessment of the agent's ability to bridge the sim-to-real gap. While some work~\cite{kadian2020sim2real,truong2023rethinking,kadian2019we} tries to predict real-world performance with the performance in simulation, the predictivity still largely relies on the specific metrics.

\subsection{Metrics}

Assessing the effectiveness of sim-to-real transfer requires the use of appropriate metrics to quantify performance discrepancies between $E_{sim}$ and $E_{real}$, as introduced in Eq.~\ref{eq:gap} about $\psi$, and such metrics are mostly relevant to the domain tasks as in Table.~\ref{tab:metric}. 

\begin{table}[!ht]
    \centering
    \resizebox{0.43\textwidth}{!}{
    {\fontsize{8}{10}\selectfont 
    \begin{tblr}{
      cell{2}{1} = {r=5}{},
      cell{7}{1} = {r=5}{},
      cell{12}{1} = {r=5}{},
      vline{2} = {1}{},
      vline{2} = {2 - 17}{},
    }
    \hline
        ~ & \textbf{Domain Metrics $\psi$} & \textbf{Relevant Works} \\ \hline
        \textbf{Robotics} & Success rate (task) & ~\cite{hofer2021sim2real, peng2018sim} \\ \hline
        ~ & Execution time & ~\cite{pettersson2005execution, nonoyama2022energy, sandha2021sim2real} \\ \hline
        ~ & Planning efficiency & ~\cite{liu2024td3, ma2022bi} \\ \hline
        ~ & Energy efficiency & ~\cite{skekala2024selected, nonoyama2022energy} \\ \hline
        ~ & Failure rate (system/task) & ~\cite{li2024optimizing, raviola2021harmonic} \\ \hline
        \textbf{Transportation} & Delay & ~\cite{wei2018intellilight, pang2024scalable} \\ \hline
        ~ & Throughput & ~\cite{wei2019colight, tsitsokas2023two} \\ \hline
        ~ & Queue length & ~\cite{wei2019survey, agarwal2024dynamic} \\ \hline
        ~ & Travel time & ~\cite{chen2020toward, gu2024pi} \\ \hline
        ~ & Pressure & ~\cite{wei2019presslight, xu2024ped} \\ \hline
        \textbf{Recom. Systems} & Click-through rate (CTR) & ~\cite{zhang2021deep, zhang2021deep} \\ \hline
        ~ & Precision & ~\cite{xiang2024integrating, khelloufi2024multimodal} \\ \hline
        ~ & Recall & ~\cite{xu2024prompting, deldjoo2024understanding} \\ \hline
        ~ & Conversion rate & ~\cite{xu2024intelligent, gao2024causal} \\ \hline
        ~ & Satisfaction score & ~\cite{min2024wisdom, wang2024deep} \\ \hline
    \end{tblr}
    }}
    \caption{Domain metrics and relevant works for Robotics, Transportation, and Recommender (Recom.) Systems.}
    \label{tab:metric}
\end{table}


Based on Eq.~\ref{eq:gap}, $G(\pi) : = \psi_s (\pi_s) - \psi_r (\pi_s)$, we can evaluate the sim-to-real gap in the dimension of $\psi$, given policies $\pi^i$ and $\pi^j$ trained from two methods $i$ and $j$, if $|G(\pi^i)| < |G(\pi^j)|$ and this difference is significant relative to the standard deviation from multiple runs, we conclude with statistical confidence that method $i$ is seen as better sim-to-real performance than $j$. 

According to different tasks, the metrics calculation varies from sparse to dense, and the indicator can also integrate more than one dimension for evaluation. To include more than one metric in a joint reflection of performance, it is feasible to define a reward function\footnote{Note: this function is just used to describe how much gain from policy executions, it is not the same as the RL training reward function.} such that: 
$R = \psi_1 + \psi _2 + ...$ (consistent numerical directions), and the gap $G (\pi)$ of the policy network $\pi$ can be written into a cumulative reward difference:
\[
\Delta R = \sum_{t=1}^{T} R_t^{\text{sim}} - \sum_{t=1}^{T} R_t^{\text{real}},
\]
where \(R_t^{\text{sim}}\) and \(R_t^{\text{real}}\) represent the rewards at time step \(t\) in simulation and real-world environments, respectively, and \(T\) is the total number of time steps (please note that for sparse settings, \(T\) is only valid for the last step). A larger \(\Delta R\) indicates greater performance degradation during transfer~\cite{polvara2020sim, da2020uncertainty}. The cumulative reward difference compares the total rewards accumulated by the agent in simulation versus real-world settings. 



\section{Open Challenges and Opportunities}\label{sec:challenge}

The field of \ours transfer in RL has been studied for years; however, several challenges persist, and new issues have emerged with the advent of large language models (LLMs) and foundation models. These challenges can be categorized into two dimensions: existing problems in \ours and new problems arising from the integration of LLMs.

\noindent\textbf{Existing Challenges in Sim-to-Real Transfer}

\textit{\underline{Simulation Fidelity and Environment Complexity}:} A major challenge is achieving simulations that are detailed enough to replicate the complexities of real-world environments. Many simplified models fail to capture the subtle interactions in physical systems~\cite{tran2023truly}, which often leads to reduced performance when transferring policies to the real world~\cite{hawasly2022perspectives}. To address this, more advanced simulators and methods need to be developed by incorporating real-world dynamic data into simulations~\cite{da2024cityflower}, which could significantly improve their accuracy, making sim-to-real transfers more reliable.


\textit{\underline{Safety and Ethical Considerations}:} Deploying reinforcement learning policies in the real world can create safety risks, especially when the policies have not been thoroughly tested against unknown situations. Addressing this requires designing comprehensive testing protocols and embedding safety constraints directly into the training process to reduce potential hazards and ensure safer deployments~\cite{chen2024trustworthy}, and another direction is developing offline-policy evaluation methods for a comprehensive understanding of policies' performance in worst cases~\cite{da2024probabilistic}, this could bring insights for pre-deployment assessment.

\noindent\textbf{Challenges Arising with Foundation Models}

\textit{\underline{Hallucination in Foundation Models}:}  Foundation models, such as large language models (LLMs), often exhibit hallucination, where they generate outputs that are factually incorrect or inconsistent with real-world data~\cite{ji2023survey}. When integrated with reinforcement learning (RL) agents, this issue becomes critical, as hallucinations can lead to suboptimal or even unsafe decision-making during task execution. Preliminary research has leveraged techniques such as grounding LLM outputs with real-world data by incorporating retrieval-augmented generation (RAG) approaches~\cite{lewis2020retrieval} and conducting LLM uncertainty quantifications to probe its real understanding of the questions~\cite{lin2023generating, da2024llm, ling2024uncertainty}, to demonstrate possible mitigation for this challenge, yet this problem is still awaiting further exploration.

\textit{\underline{Scalability and Computational Resources}:}
Large language models' inference time is proportional to their abilities, the more powerful an LLM, the larger the model parameter size, and the slower the inference speed~\cite{zhou2024survey}. When these models are combined with reinforcement learning frameworks, the demand for resources becomes even greater, it is especially costly to involve the LLM inference in each of the RL training steps. Research into task-specific distillation for LLMs~\cite{latif2023knowledge}, lighter but efficient foundation model designs can help reduce these costs~\cite{luo2024cheap}, enabling practical implementations of language model-enhanced reinforcement learning~\cite{chen2024integrating}.

In conclusion, the opportunities co-exist with those challenges, advancing the field of RL, particularly in enhancing the effectiveness and safety of sim-to-real transfers could substantially benefit from large foundation models if the above critical challenges can be handled properly.





\section{Conclusion}\label{sec:conclusion}
This paper provides a comprehensive survey of the sim-to-rea problem in Reinforcement Learning (RL), categorizing solutions within the Markov Decision Process (MDP) framework and highlighting advancements from traditional techniques to those empowered by foundation models. While significant progress has been made in mitigating the \ours gap through domain randomization, domain adaptation, and reward shaping, etc., challenges in robustness, scalability, and evaluation remain critical. By summarizing key techniques, domain-specific insights, and evaluation methods, this work serves as a foundational reference for researchers aiming to address the complexities of \ours transfer and provides resources for future innovations in realistic RL deployment.

\begin{table}[htbp]
\centering
\caption{The involved notation and explanation}
\label{tab:notation}
\begin{tabular}{@{}ll@{}}
\toprule
\textbf{Notation} & \textbf{Explanation} \\ \midrule
$M = (S, A, T, R, \gamma)$ & Markov Decision Process (MDP) \\
$S$ & State space \\
$A$ & Action space \\
$T$ & Transition function, i.e., $T(s_{t+1}|s_t,a_t)$ \\
$R$ & Reward function \\
$\gamma$ & Discount factor \\
$\pi$ & Decision policy mapping states to actions \\
$J(\pi)$ & Expected return (accumulated reward) \\
$\mu_0$ & Initial state distribution \\
$E_{sim}$ & Simulation environment \\
$E_{real}$ & Real-world environment \\
$G(\pi)$ & Sim2Real gap, $G(\pi)=\psi_s(\pi_{sim})-\psi_r(\pi_{sim})$ \\
$\psi$ & General evaluation metric \\
$\Delta_{\text{perception}}$ & Mismatch in observations \\
$\Delta S$ & Mismatch in feature representations of states \\
$\Delta A$ & Action space gap \\
$\Delta_{\text{system}}$ & System dynamics gap \\
$R_{sim}$ & Reward obtained in simulation \\
$R_{real}$ & Reward obtained in the real world \\
$\Delta R$ & Sim2Real evaluation gap\\
\bottomrule
\end{tabular}
\end{table}




\bibliographystyle{ACM-Reference-Format}
\bibliography{sample-base}











\end{document}